\documentclass[10pt,twocolumn,letterpaper]{article}

\usepackage{cvpr}
\usepackage{times}
\usepackage{epsfig}
\usepackage{graphicx}
\usepackage{amsmath}
\usepackage{amssymb}
\usepackage{enumitem}
\usepackage{rotating}
\usepackage{amsfonts}
\usepackage{empheq}
\usepackage{framed, color}
\usepackage{xcolor}
\usepackage{graphics}
\usepackage{algorithm}
\usepackage{algorithmic}
\usepackage{stmaryrd}
\usepackage[mathscr]{eucal}
\usepackage{lineno}
\usepackage{color}
\usepackage{filecontents}
\usepackage{multirow}
\usepackage{verbatim} % begin{comment} end{comment}
\usepackage{tabularx}
\usepackage{setspace}
\usepackage{textcomp,booktabs}
\usepackage{scalerel}
\usepackage{lipsum}
\usepackage{bm}
\usepackage{arydshln}
\usepackage{relsize}
\usepackage{times}
\usepackage{epsfig}
\usepackage{graphicx}
\usepackage{amsmath}
\usepackage{amssymb}
\usepackage{enumitem}
\usepackage{rotating}
\usepackage{amsfonts}
\usepackage{empheq}
\usepackage{framed, color}
\usepackage{xcolor}
\usepackage{graphics}
\usepackage{algorithm}
\usepackage{algorithmic}
\usepackage{stmaryrd}
\usepackage[mathscr]{eucal}
\usepackage{lineno}
\usepackage{color}
\usepackage{filecontents}
\usepackage{multirow}
\usepackage{verbatim} % begin{comment} end{comment}
\usepackage{tabularx}
\usepackage{setspace}
\usepackage{textcomp,booktabs}
\usepackage{scalerel}
\usepackage{lipsum}
\usepackage{bm}
\usepackage{arydshln}
\usepackage{relsize}

\definecolor{lightgray}{gray}{0.9}

\definecolor{darkgreen}{RGB}{0,112,0}
\newcommand*{\mathcolor}{}
\def\mathcolor#1#{\mathcoloraux{#1}} 
\newcommand*{\mathcoloraux}[3]{%
  \protect\leavevmode
  \begingroup
    \color#1{#2}#3%
  \endgroup
}
\usepackage{makecell}

\def\0{{\bf 0}}
\def\1{{\bf 1}}

\def\eg{\emph{e.g.}}
\def\ie{\emph{i.e.}}

\newcommand{\real}{\rm I\!R}

\usepackage[pagebackref=true,breaklinks=true,colorlinks,bookmarks=false]{hyperref}
\hypersetup{
colorlinks=true,
linkcolor=red,
filecolor=magenta,      
urlcolor=cyan,
}   

\def\cvprPaperID{2} % *** Enter the wacv Paper ID here
\def\confName{CVPR}
\def\confYear{2022}

\begin{document}

%%%%%%%%% TITLE
\title{\textit{Tragedy Plus Time}: Capturing Unintended Human Activities from \\ Weakly-labeled Videos}
% \title{\textit{Tragedy Plus Time}: Teleological Action Understanding in \\  Unintentional Human Activities}
\author{\large	Arnav Chakravarthy,  \ Zhiyuan Fang, \  Yezhou Yang \\  Arizona State University  \\
\texttt{\small achakr37@asu.edu, zy.fang@asu.edu, yz.yang\,@asu.edu} \\
}

% \pagestyle{plain}

% \twocolumn[{%
% \renewcommand\twocolumn[1][]{#1}%
\maketitle

% \begin{center}
% \includegraphics[width=.95\textwidth]{figure/intro-pic.pdf}
% \vspace{-4mm}
% \captionof{figure}{
% \small We can see how state of the art action recognition models trained on traditional datasets view unintentional action scenes. Even though this scene involves a man falling off his bike, the man's ultimate goal was to perform a stunt. The green lines indicate the regions of the video which indicate the man's goal and the red lines indicates the regions where this goal was disrupted.
% }
% \label{fig:car-crash}
% \end{center}}]

% \thispagestyle{empty}
 
%%%%%%%%% ABSTRACT
\begin{abstract}
     In videos that contain actions performed unintentionally, agents do not achieve their desired goals. In such videos, it is challenging for computer vision systems to understand high-level concepts such as goal-directed behavior, an ability present in humans from a very early age. 
     Inculcating this ability in artificially intelligent agents would make them better social learners by allowing them to evaluate human action under a teleological lens. 
     To validate this ability of deep learning models to perform this task, we curate the W-Oops dataset, built upon the Oops dataset~\cite{Epstein_2020_CVPR}.
     W-Oops consists of 2,100 unintentional human action videos, with 44 goal-directed and 30 unintentional video-level activity labels collected through human annotations.
     Due to the expensive segment annotation procedure, we propose a weakly supervised algorithm for localizing the goal-directed as well as unintentional temporal regions in the video leveraging solely video-level labels. In particular, we employ an attention mechanism based strategy that predicts the temporal regions which contributes the most to a classification task. Meanwhile, our designed overlap regularization allows the model to focus on distinct portions of the video for inferring the goal-directed and unintentional activity, while guaranteeing their temporal ordering. Extensive quantitative experiments verify the validity of our localization method.
     We further conduct a video captioning experiment which demonstrates that the proposed localization module does indeed assist teleological action understanding. Project website can be found at: \href{https://asu-apg.github.io/TragedyPlusTime}{\url{https://asu-apg.github.io/TragedyPlusTime}}.
\end{abstract}

%%%%%%%%% BODY TEXT
\section{Introduction}
 
\begin{figure}[t!]
\centering
\includegraphics[width=.46\textwidth]{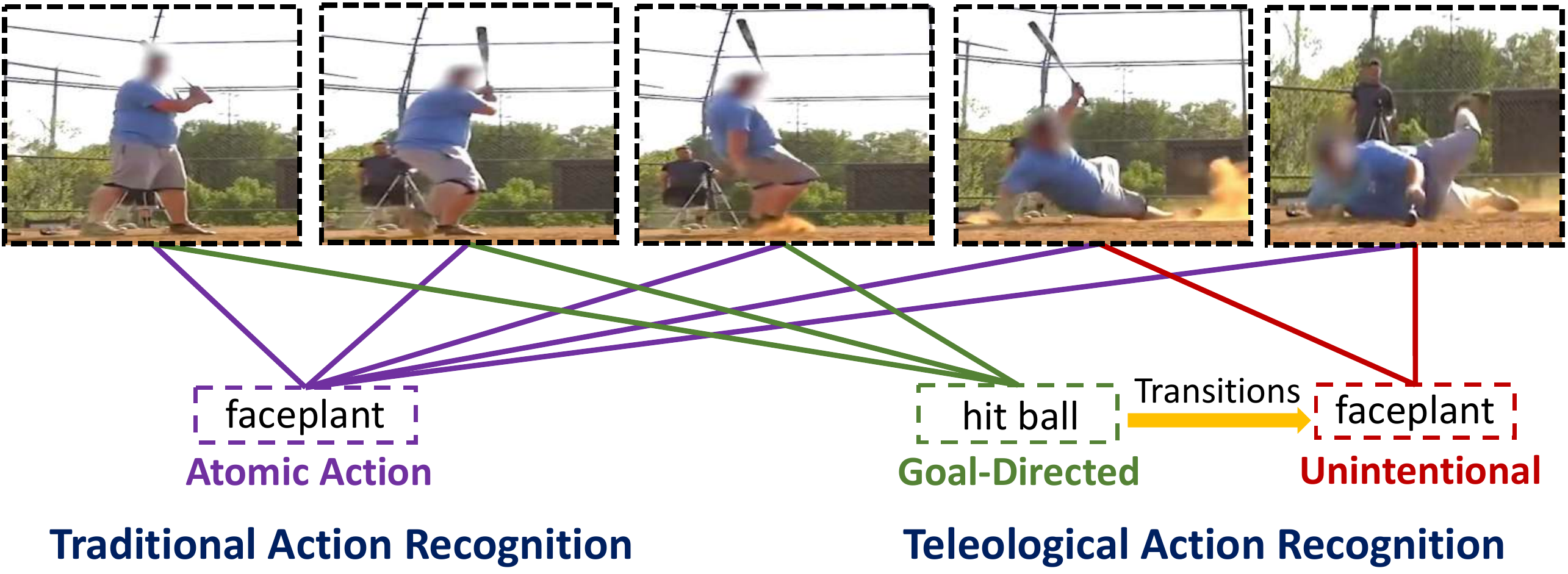}
\captionof{figure}{
\small State-of-the-art action recognition models trained on traditional video activity datasets view an unintentional action scene as an atomic action. Although this scene involves a man falling on his face, the man's ultimate goal was to hit the ball. Green lines indicate the regions of the video which indicate the man's goal, red lines indicate the regions where the action deviates from the goal, and purple lines indicate the region the action recognition model focuses on.
}
\vspace{-5mm}
\label{fig:car-crash}
\end{figure}
Traditional video action recognition \cite{ar1,ar2,ar3,ar4,ar5,carreira2017quo,tran2018closer} focuses on predicting only atomic actions present on the surface appearance of a video. On the other hand, teleological understanding of actions entails understanding the underlying goal of actions and why it was performed. These goals can be easily inferred from intentional actions as they are directly defined by their outcome. However, many actions do end up in unintended results where the goal of the action is partially or never achieved. As shown in Fig.~\ref{fig:car-crash}, an agent tries to hit the ball with a bat, but ends up landing on his face, hence not being able to achieve his goal of hitting the ball. State-of-the-art action recognition models are all trained with vewing the whole scene as ``\texttt{faceplant}'' without paying attention to the goal-directed behavior which was to ``\texttt{hit the ball}''.

Teleological action understanding provides the invaluable ability to explain and justify an action, as well as learn from mistakes in the case the goal was not achieved \cite{csibra2007obsessed}. 
For fine-grained understanding of unintentional actions, it is important to know the goal of the action, why was it not fulfilled, and when (in time) did the action start transitioning away from its goal. These abilities are present in humans from a very early age ~\cite{brandone2009you, sommerville2005pulling, woodward2001infants,gergely2003teleological,csibra2008goal}. However, this is a challenging task for deep learning models since it requires the model to understand high level concepts such as goal-directed behavior from unintentional actions which would not be possible without penetrating deeper than the surface appearance of the action~\cite{brandone2009you}.
% which is not directly visible on the surface appearance of the video, as shown in Fig.\ref{fig:car-crash}.
There are few previous works which have taken initial steps towards teleological action understanding. \cite{Epstein_2020_CVPR} builds a dataset rich in unintentional human actions, along with single point transition times manually labeled by human annotators which separate the intentional and unintentional regions of the video. They also train models on classification and localization tasks. However, it does not contain well defined classes for the goal-directed action or why this goal gets disrupted. \cite{Synakowski_2021} focuses on predicting whether an activity was intentional or unintentional, but not the understanding of the underlying goal of an unintentional action.
Previous efforts \cite{fang2020video2commonsense,lei2020more,zellers2019vcr} have tried to speculate about all possible effects of actions but do not explore which effects are undesirable.

In order to build a model which is capable of capturing the unintended activity, it is crucial to first build a dataset containing goal-directed actions and why they get disrupted Additionally, in order to localize their respective regions in time, one may manually label the transition point as in~\cite{Epstein_2020_CVPR} and fully supervise the training. However, these annotations are prohibitively expensive to collect and suffer from human error and bias. Previous works such as \cite{nguyen2018weakly,wsal2,wsal3, wsal4, wsal5, wsal6, wsal7}, which focus on segmenting atomic action scenes from untrimmed videos tackle this problem of expensive manual labelling by training a model in a weakly supervised manner using only video level action labels. Though this task differs from our task (which involves
% since we do not deal with segmenting atomic actions from untrimmed videos but 
separating the goal-directed action from the region it starts deviating from its goal), it still provides encouragement to solve our task in a weakly supervised manner. 
We bring \textbf{W}eakly Augmented \textbf{Oops} (W-Oops), an augmented human activities dataset which contains ``\textit{fail}'' videos, building upon Oops \cite{Epstein_2020_CVPR} but also contains high quality video-level annotations which describe the goal-directed as well as unintentional actions in the video.  
% As an example, Figure~\ref{fig:car-crash} shows that the relation between the agent and his goal-directed action is to ``\textit{perform a stunt}'', but this goal gets disrupted and the agent results in ``\textit{falling off the bike }''.
We further develop an
% baseline 
algorithm which allows the model to attend to contextualized visual cues to localize these regions as well as associate them with their respective goal-directed/unintentional class label, leveraging solely video-level labels. Our proposed learning schema includes an encoder to encode the joint representation of the goal-directed and unintentional action in the video as well as temporal attention modules which help the model focus on the respective regions of interest in the video. We also introduce a novel optimization target known as Overlap Regularization which allows the model to pay attention to distinct parts of the video for inferring both types of actions while ensuring their temporal ordering. Finally, we use class-agnostic (bottom-up) as well as class-specific (top-down) attention mechanisms to localize both types of actions. Additionally, we conduct a video captioning experiment leveraging our localization module to demonstrate it's teleological ablity.

To summarize our contributions:
\begin{enumerate}[nosep,noitemsep]
 \item We curate W-Oops, an augmented video dataset, built upon Oops~\cite{Epstein_2020_CVPR}, containing high quality video level labels for the goal-directed as well as unintentional action. To the best of our knowledge, we are the first to make a step towards such fine-grained understanding of unintentional actions.
 \item We propose an attention mechanism based method that highlights relevant temporal regions of the video important to a classification task when inferring the goal-directed and unintentional action while also ensuring their temporal ordering.
%  method, which incorporates attention mechanism to focus on relevant temporal regions of the video important to a classification task, by enforcing the model to pay attention to distinct parts of the video when inferring the goal-directed and unintentional action while also ensuring their temporal ordering.
 \item Finally, we provide in-depth and comprehensive experimental analysis, which validates the effectiveness of our method compared to competitive weakly supervised action localization methods on W-Oops. Additionally, we demonstrate the teleological ability of our localization module through a video captioning experiment.
 \end{enumerate}
% \jacob{ P5: Summarize our contribution, which as previously stated, plus the results we got. Which roughly includes: 1. We curated W-Oops, a novel and new video datasets consisting unintended human activities with video label action labels etc. 2. We propose a method, which .... 3. We provide in-depth and comprehensive experimental analysis, and shows that ....}

% Our contributions are two-fold :
% \begin{itemize}
% \item Providing a new dataset
% \end{itemize}

% \begin{figure*}[t!]
%     \centering
%     \includegraphics[width=.85\textwidth]{figure/fail_sample.png}
%     \caption{We can see how an action performed by an agent can transition from it's original goal. The agent's goal is to skateboard, but he ultimately loses balance and falls off the skateboard.}
%     \label{fig:fail-sample}
%     %\vspace{-1mm}
% \end{figure*}

\section{Related Work}
% \jacob{Current citations need to be increased, at least cite 70 papers: add more references for WSAL, and attach more intention related papers.}
\noindent\textbf{Intent Recognition from Visual Input.} There has been an increasing research focus on intent recognition of agents in videos.
\cite{where-why} proposes a hierarchical model to predict the intention, as well as the attention of an agent's eye gaze from a RGB-D video. \cite{vondrick2016predicting} focuses on predicting the action, motivation and scenes from an image by using a third order factor graph built using text.
\cite{Synakowski_2021} proposes an unsupervised algorithm to discriminate between an intentional and unintentional action performed by an agent. \cite{Epstein_2020_CVPR} too focuses on discriminating between an intentional and unintentional action, as well as predicting the point in an unintentional video when the action deviates from it's original goal, but does this in a supervised manner. Our work differs from these as we focus on discriminating between the different goal-directed and unintentional action categories in unintentional videos, as well as localizing these action regions in a weakly supervised manner.
Action anticipation can also be relevant to predicting an unintentional action or the onset of it. \cite{rolling-unrolling,sadegh2017encouraging,miech2019leveraging, ryoo2011early,hoai2012max} focus on forecasting an event or action based on a small snippet of a video.
\cite{rep-pred, back-to-future} focus on self supervised learning approaches to predict future action representation using unlabeled videos. 
\cite{fang2019intention,rasouli2020pedestrian,intent-urban} focus on predicting a pedestrian's intent to cross the road using the Joint Attention for Autonomous Driving (JAAD) dataset \cite{JAAD}. Our work differs from this as it not only focuses on the past and not on predicting the future, but is also generalized to more diverse environmental settings.

\noindent\textbf{Weakly Supervised Action localization (WSAL)}  has been drawing increasing attention due to the expensive manual labelling process involved in a fully supervised setting. Previous efforts involve localizing action regions in an untrimmed video by training a model using only video level action labels~\cite{sun2015temporal,shou2018autoloc,nguyen2018weakly,paul2018w, buch2017sst,kalogeiton2017action,weinzaepfel2015learning,shou2017cdc,tran2012max,shou2016temporal}, or sentences~\cite{Mithun_2019_CVPR,fang2020weak,chen2020look,fang2018weakly,song2020weakly,fang2019modularized,ma2020vlanet}. In particular, STPN~\cite{nguyen2018weakly} trains a classification model using sum of features weighted by their class-agnostic attention weights, which it learns using a sparsity loss on the attention weights. It then performs the localization by using both the classification activation as well as these class-agnostic weights, thresholding them to select action locations.  WTALC~\cite{wsal6} forces the foreground action features from the same action class to be similar and the background features pertaining to an action class to be dissimilar from its foreground feature, and finally localizes the action by thresholding the classification activation. A2CL-PT~\cite{wsal2} uses foreground and background features to form triplets and apply the Angular Triplet Center Loss \cite{li2019angular} to separate the foreground and background features, as well as use an adversarial branch in order to find supplementary activities from non-localized parts of the video.
DGAM~\cite{wsal3} propose to separate action frames from context frames by modeling the frame representation conditioned on the bottom-up attention.
TSCN~\cite{wsal4} fuse the attention sequences from the RGB and optical flow stream and use it as pseudo ground truth to supervise the training.\

\begin{figure*}[t!]
    \centering
    \includegraphics[width=.90\textwidth]{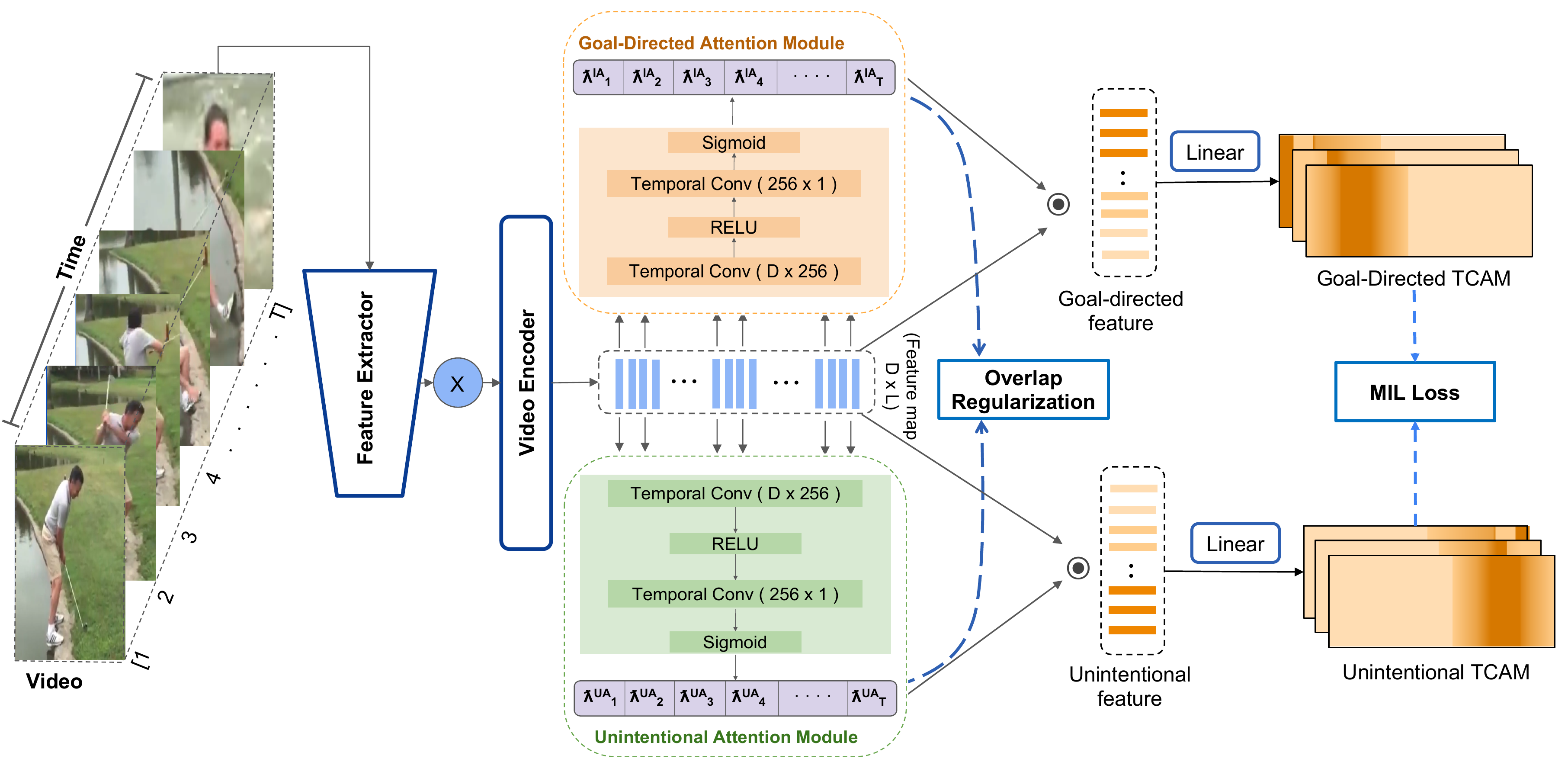} 
    \vspace{-2mm}
    \caption{\small Illustration of our overall architecture. A backbone feature extractor is used to convert raw videos into features, \ie , $\mathbf{X}$ and is kept frozen throughout the training process. $\mathbf{X}$ is then passed to a video encoder which can be either a GRU~\cite{chung2014empirical} or a Transformer Encoder~\cite{NIPS2017_3f5ee243}, to extract high level features $\mathbf{O}$. The two attention modules use $\mathbf{O}$ to predict the bottom-up attention weights $\lambda^\text{IA}$ and $\lambda^\text{UA}$ for the goal-directed and unintentional action respectively, which are used for the Overlap Regularization. We calculate the goal-directed, \ie, $\mathbf{O}^\text{IA}$ and unintentional feature, \ie, $\mathbf{O}^\text{UA}$ by computing a dot product between $\mathbf{O}$ and their respective bottom-up attention weights. Finally we pass the goal-directed and unintentional feature through weight-shared linear layers to extract their respective TCAMs $\mathbf{C}^\text{IA}$ and $\mathbf{C}^\text{UA}$. These TCAMs are used for the MIL Loss.   }
    \label{fig:architecture}
    \vspace{-3mm}
\end{figure*}

\section{Methodology}

We intend to identify the goal-directed and unintended human activities, as well as their corresponding moment of occurrence from an unintentional video in a weakly-supervised manner. To be specific, given the video $\mathcal{V}$ and its categorical labels representing the goal-directed activity, $y^\text{IA}$, and the unintended activity, $y^\text{UA}$, we expect the model to predict the triplets $\langle s^\text{IA}, e^\text{IA}, c^\text{IA}\rangle$ and $\langle s^\text{UA}, e^\text{UA}, c^\text{UA}\rangle$, containing the starting point, end point and action class associated with this segment by leveraging only the video-level annotations as weak supervision. We formulate this challenge as a weakly supervised action localization (WSAL) task, and address it using an attention mechanism based approach. We start this section by providing an overview of our model, followed by the details of formulations and our proposed learning objective.

% \jacob{for simple and easier understanding, avoid say we have a set of goal-directed labels. Just describe as if we only have one label per video. And clarify this details in experiments.}

% \jacob{Why do we need to predict out the probability p? is this symbol used anywhere in later content?}

% More formally, consider that we have a video of length $l$, with the goal-directed activity label set represented by $\textbf{\textit{$A_g$}}=\{{{a}^g}\}$ and unintended label activity set $\textbf{\textit{$A_u$}}=\{{{a}^u}\}$. During test time, given a video \textit{V}, we need to predict two sets $v_{det}^g = \{(s_j^g,e_j^g,c_j^g,p_j^g)\}$ and $v_{det}^u=\{(s_j^u,e_j^u,c_j^u,p_j^u)\}$ where $v_{det}^g$ and $v_{det}^u$ are detection sets for the goal-directed and unintentional activity respectively. $s_j$, $e_j$ are the start and end times of the $j^{th}$ detection, $c_j$ represents its predicted activity category with confidence $p_j$.

\subsection{Task Formulation}
To encode the videos, pre-trained 3D neural networks are exploited to extract a set of clip-level representations $\mathbf{X}$.
We find that in order to encode the goal-directed and unintentional features from the video, directly using static features is not sufficient. Hence, we encode the clip embedding by an encoder network $\mathcal{F}$, which outputs a joint representation for the goal-directed and unintentional action:
% that captures their temporal relations:
%\begin{equation}
%    \small 
%    \mathbf{O} = \mathcal{F}(\mathbf{X}),
%    % \phi_i = \mathcal{F}(X_i),
%\end{equation}
$
    \mathbf{O} = \mathcal{F}(\mathbf{X}),
$
where $\mathbf{O}\in\mathbb{R}^{l \times d}$ denotes the representations in $d$ dimensions for $l$ clips. Here, encoder network $\mathcal{F}$ can either be a bidirectional Gated Recurrent Unit or a Transformer Encoder~\cite{NIPS2017_3f5ee243}. On this basis, we introduce two bottom-up attention modules, which outputs the temporal attention weights that reflect the temporal importance of clip representations for the goal-directed/unintentional activity respectively. This is achieved by training the model with a classification loss, \eg, multiple instance learning loss. Note that these attention weights are agnostic to the specific action, and are used to identify generic regions of interest. A stack of 1-D Convolution layers with RELU activation between layers, followed by a Sigmoid function is used to obtain the attention weights $\lambda^\text{IA}$, $\lambda^\text{UA} \in \mathbb{R}^{l}$ with a scale between 0 and 1.\

% The result is attention weights for the goal-directed action \ie~$\lambda^{IA}\in \real^{1\times{l_i}}$ as well as unintentional action \ie~$\lambda^{UA} \in \real^{1\times{l_i}}$ .Note that this module is not weight shared across the goal-directed and unintentional region attention calculation.\par
\indent In order to obtain goal-directed and unintentional features, we compute a dot product between the joint representation $\mathbf{O}$ and each of the bottom-up attention weights $\lambda^\text{IA}$ and $\lambda^\text{UA}$. These features would represent those parts of the joint representation $\mathbf{O}$ which correspond to the goal-directed and unintentional region respectively. Formally,
\begin{align}
    \small 
        \mathbf{O}^\text{IA} &= \mathbf{O} \cdot \lambda^\text{IA}, & \mathbf{O}^\text{UA} &= \mathbf{O} \cdot \lambda^\text{UA}.
    \end{align}

We then compute Temporal Class Activation Maps (TCAM)~\cite{nguyen2018weakly}, $\mathbf{C}^\text{IA} \in \mathbb{R}^{l\times N_\text{IA}}$, $\mathbf{C}^\text{UA} \in \mathbb{R}^{l\times N_\text{UA}}$ for the goal-directed as well as unintentional actions, with $N_\text{IA}$ and $N_\text{UA}$ corresponding to the number of goal-directed and unintentional classes, by employing two weight-sharing linear transformation layer on $\mathbf{O}^\text{IA}$ and $\mathbf{O}^\text{UA}$ respectively.  These are one dimensional class-specific activations that signify classification scores over time for both the types of actions for each segment (as illustrated in Fig.~\ref{fig:architecture}). 
% Note that the TCAMs are class-specific while the bottom-up attention weights \textit{i.e} $\lambda$ are class-agnostic.
% The TCAM outputs $\mathbf{C}^\text{IA}$ and $\mathbf{C}^\text{IA}$ in video $\mathcal{V}$ can be retrieved by feeding the $\mathbf{O}^\text{IA}$ and $\mathbf{O}^\text{UA}$ to a weight-shared linear transformation layer, as shown in Fig. \ref{fig:architecture}
% \begin{equation} \label{tcam-eqn}
%     \begin{aligned}
%     % \mathbf{T}^\text{IA}_c = \mathbf{W}^\text{IA}_c\cdot\text{RELU}(\phi_i^{IA}) \\
%     \C^{UA}(i,c) = W_{fc}^{UA}(c)\cdot{Relu(\phi_i^{UA})} \oplus b_{fc}^{UA}(c).
%     \end{aligned}
% \end{equation}
These class-specific distributions, along with the class-agnostic distributions are used to predict the triplets $\langle s^\text{IA}, e^\text{IA}, c^\text{IA}\rangle$ and $\langle s^\text{UA}, e^\text{UA}, c^\text{UA} \rangle$ associated with the goal-directed and unintentional activities respectively.
\subsection{Video Encoder}
In order to learn a joint representation for inferring the goal-directed and unintentional actions, we use a Bidirectional Gated Recurrent Unit~\cite{chung2014empirical} as the video encoder. 3D-CNN architectures like R(2+1)D~\cite{tran2018closer} and I3D~\cite{carreira2017quo} capture very short clip level information. However, capturing information which helps discriminate between the goal-directed region and an unintentional region requires longer temporal context which can be modeled by a GRU. Specifically, our GRU consists of a reset gate $r$ which controls how much importance to give the previous hidden state $h^{t-1}$ in order to calculate the current hidden state $h^{t}$, and an update gate $u$ which determines how much of the previous hidden state $h^{t-1}$ should be carried on to the current hidden state $h^{t}$. Given the backbone feature $\mathbf{X}$, we compute the hidden state at each time-step $t$ using the following equations:

\begin{equation}
    \small 
    \begin{aligned}
        z^{t} &= \sigma(\mathbf{W}
^{z} \mathbf{X}^{t} + \mathbf{U}^{z} h^{t-1)}) & \text{Update Gate} \\
        r^{t} &= \sigma(\mathbf{W}
^{r} \mathbf{X}^{t} + \mathbf{U}^{r} h^{t-1}) & \text{Reset Gate} \\
        \tilde{h}^{t} &= \tanh(r^{t}\cdot \mathbf{U} h^{t-1} + \mathbf{W} \mathbf{X}^{t}) & \text{New Memory} \\
        h^{t} &= (1-z^{t})\cdot\tilde{h}^{t} + z^{t} \cdot h^{t-1}, & \text{Hidden State} \\
    \end{aligned}
\end{equation}
where $\mathbf{U}$ and $\mathbf{W}$ correspond to learnable parameters of this module.
In order to capture the forward information flow $\overrightarrow{h^{(t)}}$ as well as backward information flow $\overleftarrow{h^{(t)}}$ we use a Bidirectional-GRU and obtain the final representation $\mathbf{O}$ by concatenating these features from the final hidden layer.
% \jacob{NOTE: it is very important to keep these equations mathematically correct or they might lead to strong rejection bc technically error. E.g., use consistent symbol to represent matrix and vector, what does U represents?avoid use parenthesis in the superscript. usually matrix is in the format of mathbf mode, and vector is in plain format. Check the above again incase I miss any thing.}

\subsection{Multiple Instance Learning Loss} \label{MIL}
Following previous works in weakly supervised action localization \cite{wsal5,wsal6,wsal2}, we use the $k$-max Multiple Instance Learning (MIL) \cite{zhou2004multi} loss function for classifying the goal-directed and unintentional activities in the video. For each video, we average out the top-$k$ elements of the TCAMs, \ie, $\mathbf{C}^\text{IA}$ and $\mathbf{C}^\text{UA}$ along the temporal axis for each class to obtain the video-level classification scores $A^\text{IA} \in \mathbb{R}^{{N_{IA}}}$ and $A^\text{UA} \in \mathbb{R}^{{N_{UA}}}$. Here, $k$ is set by \big[$\frac{l}{s}$\big] where $s$ is a hyper-parameter that regulates the number of clips to consider when making the classification. We then apply a softmax function over class scores, in order to obtain a probability mass function (pmf) over the goal-directed as well as unintentional classes, \ie, $p^\text{IA}$ and $p^\text{UA}$.
Let $y^\text{IA}$ and $y^\text{UA}$ be the ground truth label vectors for a video. We then $l_1$-normalize them to obtain ground-truth pmfs $q^\text{IA}$ and $q^\text{UA}$. Finally we conduct cross entropy between these two.

\begin{equation}
\small 
\begin{aligned}
\mathcal{L}_{cls}^\text{IA} &= \dfrac{1}{N}\sum_{i=1}^N \sum_{j=1}^{N_\text{IA}} -q_i^\text{IA}(j)\log\big(p_i^\text{IA}(j)\big) \\
\mathcal{L}_{cls}^\text{UA} &= \dfrac{1}{N}\sum_{i=1}^N \sum_{j=1}^{N_\text{UA}} -q_i^\text{UA}(j)\log\big(p_i^\text{UA}(j)\big) ,\\
\end{aligned}
\end{equation} 
where $N$ corresponds to the total number of training videos, and the final loss is the combination of them: $\mathcal{L}_{cls} = \mathcal{L}_{cls}^\text{IA} + \mathcal{L}_{cls}^\text{UA}$.
% \jacob{for above equation, just use one activity label per video to avoid vague explanation. At that case, does q represents a scalar or vector? if its a tensor then the cross entropy loss should adjusted accordingly. Make sure the notation is consistent with my previous section notations!!! E.g., make sure matrix is in mathbf as previous, make superscript IA in text mode.}

% \begin{figure*}[t!]
% \centering
%     \includegraphics[width=.95\textwidth]{figure/action_dist.pdf} 
%     \vspace{-4mm}
%     \includegraphics[width=.95\textwidth]{figure/len-seg-dist.pdf} 
%     % \caption{\small Distribution over the goal-directed and unintentional actions. All action classes are not shown due to space constraints. Please refer to the appendix in order to view the full class list.}
%     % \label{fig:class-dist}
%     \vspace{1mm}
%     \caption{\small Top: Distribution over the goal-directed and unintentional actions (partially shown). Lower Left: Distribution over goal-directed and unintentional segment lengths (normalized by the video length). Lower Right: Distribution over the entire video length. }
%     \vspace{-3mm}
%     \label{fig:len-dist}
% \end{figure*}
% \vspace{-9mm}

\subsection{Overlap Regularization} \label{overlap}
Let $\lambda^\text{IA}_t$, $\lambda^\text{UA}_t \in [0,1]~{\forall}~t \in [1, l]$ be the bottom-up attention weights for the goal-directed actions and unintentional action respectively, obtained from the respective attention modules. $\lambda_t$ signifies the temporal attention weight for a clip $t$.
During training, a trivial solution which could be learned by the model is to pay attention to the entire video when inferring the goal-directed and unintentional action, \ie,  $\lambda^{IA}_{t},\lambda^{UA}_{t} = 1~{\forall}~t \in [1, l]$, though these actions take place at two distinct sections of the video.
Simply applying the MIL loss cannot guarantee that such distinctions can be learnt from the data as shown in Section \ref{ablation}.
We solve this problem by appending an additional regularization term on the overlap of these attention weights:
\begin{equation}
\!
\begin{aligned}
\small 
\mathcal{L}_{IA} &= \texttt{max}\Big(0, \dfrac{\sum_r^{N_{T^\text{UA}}}\lambda_{T_r^\text{UA}}^{IA}}{N_{T^\text{UA}}} - \dfrac{l}{p}) \\
\mathcal{L}_\text{UA} &= \texttt{max}\Big(0, \dfrac{\sum_r^{N_{T^\text{IA}}}\lambda_{T_r^\text{IA}}^\text{UA}}{N_{T^\text{IA}}} - \dfrac{l}{p}) \\
&\mathcal{L}_{overlap} = \mathcal{L}_\text{IA} + \mathcal{L}_\text{UA}, 
\end{aligned}
\end{equation}
where ${T^\text{IA}}$ and ${T^\text{UA}}$ are the set of temporal indices of the bottom-up goal-directed and unintentional attention weights at which they are more than a predefined threshold. $N_{T^\text{IA}}$ and $N_{T^\text{UA}}$ are the lengths of the sets of activated temporal indices. $p$ is a design parameter which controls the amount of allowed overlap between the attention maps. Lower the value of $p$, lower the penalization of overlaps.\
\begin{table*}[h!]
{
\setlength{\tabcolsep}{7pt} % Default value: 6pt
\renewcommand{\arraystretch}{1.02} % Default value: 1
\centering
{\small
 \begin{tabular}{c c c | c c c c c c c c c | c} 
%  \toprule \\ [-3.4ex]
\multirow{2}{*}{\textbf{Model}} & \multirow{2}{*}{\textbf{Feature}} & \multirow{2}{*}{\textbf{Segment}} & \multicolumn{10}{c}{\textbf{mAP @ IoU}}\\
 & & & 0.1 & 0.2 & 0.3 & 0.4 & 0.5 & 0.6 & 0.7 & 0.8 & 0.9 & {AVG.}\\ [0.2ex] 
\hline
\hline
     \multirow{2}{*}{STPN~\cite{nguyen2018weakly}} & \multirow{2}{*}{R(2+1)D} 
     & \text{Goal} & 44.9 & 41.7 & 33.0 & 25.7 & 18.3 & 10.0 & 5.0 & 3.7 & 1.2 & 20.4\\
     [-0.4ex]
     & & \text{UnInt} & 30.9 & 26.6 & 21.8 & 15.7 & 9.9 & 5.2 & 1.8 & 1.00 & 0.1 & 12.5 \\[0.4ex] 
     
     \multirow{2}{*}{WTALC~\cite{wsal6}} & \multirow{2}{*}{R(2+1)D}
     & \text{Goal} & 45.1 & 41.8 & 36.1 & 28.9 & 22.8 & 15.9 & 10.4 & 8.1 & 2.0 & 23.5\\[-0.4ex]
     & & \text{UnInt} & 25.5 & 21.2 & 15.3 & 12.6 & 7.7 & 4.3 & 2.3 & 1.0 & 0.5 & 10.1\\[0.4ex] 
     \multirow{2}{*}{A2CL-PT~\cite{wsal2}} & \multirow{2}{*}{R(2+1)D}
     & \text{Goal} & 41.1 & 38.8 & 34.3 & 28.4 & 23.9 & 16.6 & 10.9 & 8.4 & 2.5 & 22.8\\[-0.4ex] 
     & & \text{UnInt} & 30.2 & 24.2 & 19.8 & 14.2 & 8.6 & 5.0 & 1.8 & 0.6 & 0.1 & 11.6\\[0.4ex] 
     
     \multirow{2}{*}{Ours} & \multirow{2}{*}{R(2+1)D}
     & \text{Goal} & 45.3 & 45.1 & 44.0 & 41.8 & 39.1 & 29.5 & 21.9 & 13.9 & 3.5 & 31.6\\[-0.4ex] 
     & & \text{UnInt} & 34.6 & 33.4 & 28.4 & 23.6 & 19.5 & 15.0 & 10.0 & 3.4 & 1.0 & 18.8\\[0.4ex] 
    %  \hline
     \multirow{2}{*}{STPN~\cite{nguyen2018weakly}} & \multirow{2}{*}{I3D}
     & \text{Goal} & 44.8 & 42.8 & 34.9 & 27.8 & 19.9 & 11.1 & 6.1 & 4.0 & 1.6 & 21.5\\[-0.4ex] 
     & & \text{UnInt} & 36.3 & 31.3 & 26.1 & 19.5 & 13.0 & 6.8 & 1.7 & 0.6 & 0.02 & 15.0\\[0.4ex] 
     \multirow{2}{*}{WTALC~\cite{wsal6}} & \multirow{2}{*}{I3D}
     & \text{Goal} & 38.8 & 36.4 & 30.4 & 26.3 & 18.6 & 13.1 & 7.2 & 4.5 & 1.8 & 19.7 \\ [-0.4ex] 
     & & \text{UnInt} & 22.9 & 18.4 & 14.2 & 11.0 & 6.8 & 3.6 & 1.2 & 0.5 & 0.1 & 8.8 \\[0.4ex] 
     \multirow{2}{*}{A2CL-PT~\cite{wsal2}} & \multirow{2}{*}{I3D}
     & \text{Goal} & 38.1 & 36.7 & 31.8 & 26.6 & 22.7 & 17.6 & 12.5 & 9.0 & 4.9 & 22.2\\[-0.4ex] 
     & & \text{UnInt} & 32.4 & 26.1 & 21.6 & 15.3 & 9.9 & 5.2 & 1.6 & 0.7 & 0.1 & 12.5\\[0.4ex] 
     \multirow{2}{*}{Ours} & \multirow{2}{*}{I3D} 
     & \text{Goal} & \textbf{51.5} & \textbf{51.3} & \textbf{49.9} & \textbf{44.9} & \textbf{41.1} & \textbf{32.5} & \textbf{24.3} & \textbf{14.4} & \textbf{5.0} & \textbf{35.0}\\[-0.4ex] 
     & & \text{UnInt} & \textbf{39.4} & \textbf{39.0} & \textbf{36.4} & \textbf{32.2} & \textbf{30.0} & \textbf{26.6} & \textbf{17.6} & \textbf{10.2} & \textbf{2.8} & \textbf{26.0} \\[0.2ex] 
%  \bottomrule
 \end{tabular}
 }
  \vspace{-2mm}
 \caption{\small Performance comparison of our model with competitive weakly supervised action localization (WSAL) models. We adjust the WSAL models by attaching two classification heads to compute two TCAMs (for the goal-directed and unintentional action). We then retrain it on our dataset (W-Oops). We can see that our model significantly outperforms the other methods. }
  \vspace{-2mm}
\label{table:1}
}
\end{table*}

\indent In the goal-directed as well as unintentional regions of the video, the attention weights should ideally be low at the borders of their respective ground truth region and high towards the center of this region.
Hence we view $\lambda_\text{IA}, \lambda_\text{UA}$ as Gaussian distributions $\mathbf{P_\text{IA}} \sim \mathcal{N}(\mu_\text{IA},\,\sigma_\text{IA}^{2})$ and $\mathbf{P_\text{UA}} \sim \mathcal{N}(\mu_\text{UA},\,\sigma_\text{UA}^{2})$. Every unintentional action begins with an agent performing a goal-directed action in order to achieve it's goal, which then gets disrupted and transitions into an unintentional action. Using this prior that a goal-directed action transitions into an unintentional action, we need to ensure $\mu_\text{IA} < \mu_\text{UA}$. We approach this by formulating the following regularization:
\begin{equation}
\!
\begin{aligned}
\small 
\mu_\text{IA} &= \dfrac{\sum_{t=1}^{l} \textit{P}^{\lambda_\text{IA}}_t\cdot{t}}{\sum_{t=1}^{l} \textit{P}^{\lambda_\text{IA}}_t}\\
\mu_\text{UA} &= \dfrac{\sum_{t=1}^{l} \textit{P}^{\lambda_\text{UA}}_t\cdot{t}}{\sum_{t=1}^{l} \textit{P}^{\lambda_\text{UA}}_t}\\
\mathcal{L}_{order} &= \texttt{max}(0, \dfrac{\mu_\text{IA} - \mu_\text{UA}}{l} + \dfrac{l}{q}),
\end{aligned}
\end{equation}
where  $\textit{P}^{\lambda_\text{IA}}$ and $\textit{P}^{\lambda_\text{UA}}$ are probability distributions obtained by applying softmax over the temporal axis of $\lambda_\text{IA}$ and $\lambda_\text{UA}$ respectively. $q$ is a design parameter that helps control the margin by which  $\mu_\text{UA}$ has to be greater than $\mu_\text{IA}$.
Our model is end-to-end trained with the overall loss as follows:
\begin{equation}
\small
\mathcal{L} = \lambda{L}_{cls} + (1-\lambda)(\mathcal{L}_{overlap} + \mathcal{L}_{order}),
\end{equation}
where $\lambda$ is the weighting hyper-parameter that controls the trade-off between MIL loss and overlap regularization.

\subsection{Classification and Localization}
After training our network, we use it to classify goal-directed and unintentional actions as well as localize the regions in which they occur. For a single video, after obtaining the pmf $p^\text{IA}$ and $p^\text{UA}$ over each of the classes, as mentioned in Section \ref{MIL}, we use mean average precision (mAP) to conduct evaluation for the classification task.
For localization of the goal-directed and unintentional regions, we consider only categories having classification scores \ie,~$A^\text{IA}$ and $A^\text{UA}$ above $0$. For each of these categories, we first scale the respective TCAM outputs to [0,1] using a Sigmoid function and weight these using the bottom-up attention weights. This can be formally expressed by:
% After obtaining the TCAMs $C^\text{IA}$ and $C^\text{UA}$ for the remaining categories,
\begin{equation}
    \small 
    \begin{aligned}
    \psi^\text{IA}(c^\text{IA}) &= \lambda_\text{IA}\cdot\texttt{Sigmoid}(C^\text{IA}(c^\text{IA})) & c^\text{IA} \in [1, N_\text{IA}], \\
    \psi^\text{UA}(c^\text{UA}) &= \lambda_\text{UA}\cdot\texttt{Sigmoid}(C^\text{UA}(c^\text{UA})) & c^\text{UA} \in [1, N_\text{UA}],
\end{aligned}
\end{equation}
where $\psi^\text{IA}(c^\text{IA})$, $\psi^\text{UA}(c^\text{UA}) \in \mathbb{R}^{l}$ are the weighted TCAMs, for the respective classes $c^\text{IA}$ and $c^\text{UA}$.
We finally threshold $\psi^\text{IA}(c^\text{IA})$ and $\psi^\text{UA}(c^\text{UA})$ to obtain the triplets $\langle s^\text{IA}, e^\text{IA}, c^\text{IA} \rangle$ and $\langle s^\text{UA}, e^\text{UA}, c^\text{UA} \rangle$. 
% \jacob{do we have experiments or section to discuss the effect of different threshold weight?}

\section{Experiments}
\subsection{Dataset \& Implementations}
\noindent\textbf{Data Preparation.} The original Oops Dataset~\cite{Epstein_2020_CVPR} consists of 20,338 videos containing unintentional human actions obtained by collating ``\textit{fail}'' videos from different users on YouTube. Amazon Mechanical Turk workers are then asked to label the time at which the video starts transitioning from the goal-directed action to the unintentional action, as well as indicate whether a video does not indicate an unintentional action.\
% After repeating this process multiple times, the authors find that humans are consistent in labelling the time of transition.\

In order to create our dataset, which is built upon the labeled portion of the Oops dataset, we follow a similar pre-processing step as in \cite{Epstein_2020_CVPR} by removing those videos that 1). Do not contain an unintentional action 2). Are More than 30 seconds which are likely to contain multiple scenes, as well as those less than 3 seconds which are not likely to contain one full scene 3). Where the transition time occurs in the initial/ending 1\% of the video, since there would not be enough context to understand the goal-directed/unintentional action respectively. Post this process, we were left with a total of about 7,800 labeled videos.

The authors of \cite{Epstein_2020_CVPR} also provide annotations in the form of natural language descriptions, which were obtained by asking Amazon Mechanical Turkers to watch the video and answer: ``\textit{what was the goal?}'' and ``\textit{what went wrong?}''. Since we want to collect a distinct set of goal-directed and unintentional actions, we followed a technique similar to the Epic Kitchens Dataset~\cite{damen2018scaling}, by extracting the verbs and associated noun using the SpaCy\footnote{\url{https://spacy.io/}} dependency parser and concatenating them to form an action. We replace all compound nouns by it's second noun: \eg, ``\textit{ride mountain bike}'' is replaced with ``\textit{ride bike}'' and so on. Due to the diversity of the worker's vocabulary, we find that the resulting actions are of low quality, with many of them having ambiguous meanings, \ie, ``\textit{fly bike}'' as well as redundant meanings. In order to overcome this, we manually go over each of these extracted actions and remove those with ambiguous meanings as well as merge the redundant ones, \ie, ``\textit{jump over fence}'' and ``\textit{jump over chair}'' into a more general ``\textit{jump over obstacle}'' category. We finally carry out a human evaluation, going through all the videos manually and ensuring the correctness of the labels, and correcting them if need be. We also give the evaluator an option to discard the video if the goal of the actor was ambiguous. We build an annotation tool in order to make this process easier (refer to the Appendix for details).
Finally, we keep a threshold of 15 for the number of videos per goal-directed and unintentional action class, discarding all classes below this threshold, as well as the videos associated with these classes. This leaves us with 44 goal-directed and 30 unintentional classes. We provide detailed statistics and analysis of the dataset in the Appendix.

% \noindent\textbf{Statistics and Analysis.}
% The final W-oops dataset contains 1,582 train samples and 526 testing samples, containing a total of 44 diverse goal-directed and 30 unintentional action classes, which can be seen in Fig.~\ref{fig:len-dist}. We have also provided the distribution of the goal-directed and unintentional segment lengths, as well as the total video lengths. It shows that the goal-directed and unintentional segment lengths are well diversified over then entire length of the video. The lengths of the video are short in general, with a majority of them ranging from 6.2 - 7.7 seconds. This makes the task of identifying these sub-regions in the video challenging.
% In our benchmark, train samples contain only video-level labels whereas the test samples contain both the video-level labels as well as the unintended activity transition points (taken from the original Oops dataset), which we use to split the video into a goal-directed and unintentional region in order to conduct evaluation.

\begin{table}[t!]
\centering
\setlength{\tabcolsep}{4.5pt} % Default value: 6pt
{\small
 \begin{tabular}{c c |c c} 
% \toprule \\ [-3.8ex]
 \textbf{Architecture} & Feature & GOAL cMAP & UNINT. cMAP \\ 
 \hline
 \hline
 Chance & - & $2.7$ & $3.3$ \\ 
%  \hline
 STPN & R(2+1)D & $44.0$ & $32.6$ \\
  WTALC & R(2+1)D & $48.5$ & $37.5$ \\
  A2CL-PT & R(2+1)D & $46.6$ & $32.6$ \\
Ours & R(2+1)D & $50.5$ & $38.4$ \\
% \hline
 STPN & I3D & $45.3$ & $37.5$ \\
 WTALC & I3D & $50.2$ & $38.2$ \\
 A2CL-PT & I3D & $48.5$ & $34.8$ \\
 Ours & I3D & $52.6$ & $41.1$\\

%  \bottomrule
 \end{tabular}
 \vspace{-1mm}
 \caption{\small Mean average precision of activity classification results using different methods. First row shows the mAP of random chance.}
 \label{table:2}
 \vspace{-4mm}
 }
\end{table}

\noindent \textbf{Implementation Details.} We extract RGB features by creating chunks of 16 consecutive and non-overlapping frames and using the I3D~\cite{carreira2017quo} as well as R(2+1)D~\cite{tran2018closer} pretrained architectures to extract clip-level features from these chunks (details provided in the Appendix). This backbone feature extractor is kept frozen throughout the entire training process. The kernel-size of all the 1-D convolutional layers for the bottom-up attention modules are set to 1. The learning rate and loss weighting function $\lambda$ is set to $10^{-3}$ and 0.8 respectively. We set the MIL loss hyper-parameter $s$ to 3. The parameters of the Overlap Regularization, $p$ and $q$, are set to 1000 and 10 respectively. Finally we set the number of layers of our bidirectional GRU to 3. Our network is implemented and trained on a machine with a single Tesla X Pascal GPU for 10,000 iterations using the Adam Optimizer \cite{kingma2014adam} with a batch size of 16.

% \jacob{Right now this section seems to have focused too much on feature extraction part, which is not quite important for our main theme. Consider to move most part of feature extraction to Appendix, and just briefly mention it. And this saves us some more space. Some other implementation details are missing, for example: learning rate, any warm-up? network details missing? number of layers and dimension of  features? Refer to other papers and make sure implementations details are fully disclosed.}

\begin{table}[t!]
\centering
\setlength{\tabcolsep}{3.4 pt} 
{\small 
 \begin{tabular}{c c c| cc c c c} 
%  \toprule \\ [-3.8ex]
\multirow{2}{*}{$\mathcal{L}_{cls}$} & \multirow{2}{*}{$\mathcal{L}_{order}$} & \multirow{2}{*}{$\mathcal{L}_{overlap}$} & \multirow{2}{*}{SEG.} & \multicolumn{4}{c}{\textbf{mAP @ IoU}} \\ 
& & & & 0.3 & 0.5 & 0.9 & AVG. \\[0.2ex] 
 \hline
  \hline
\multirow{2}{*}{\checkmark} & \multirow{2}{*}{-} & \multirow{2}{*}{-} 
& \text{Goal} & $34.7$ & $17.6$ & $0.9$ & $21.2$ \\[-0.4ex] 
& & & \text{UnInt} & $31.1$ & $14.4$ & $0.1$ & $17.4$ \\[0.4ex] 
\multirow{2}{*}{\checkmark} & \multirow{2}{*}{\checkmark} & \multirow{2}{*}{-}
& \text{Goal} & $46.3$ & $35.2$ & $2.7$ & $30.1$ \\[-0.4ex] 
& & & \text{UnInt} & $31.7$ & $17.9$ & $0.7$ & $19.0$ \\[0.4ex] 
\multirow{2}{*}{\checkmark} & \multirow{2}{*}{\checkmark} & \multirow{2}{*}{\checkmark}
& \text{Goal} & $49.9$ & $41.1$ & $5.0$ & $35.0$ \\[-0.4ex] 
& & & \text{UnInt} & $36.4$ & $30.0$ & $2.8$ & $26.0$ \\[0.4ex] 
%  \bottomrule
 \end{tabular}
 \vspace{-1mm}
 \caption{\small Ablation study on contributions of different losses in our model.}
 \label{table:4}
  \vspace{-4mm}
 }
\end{table}

\begin{figure*}[t!]
    \centering
    \includegraphics[width=.95\textwidth]{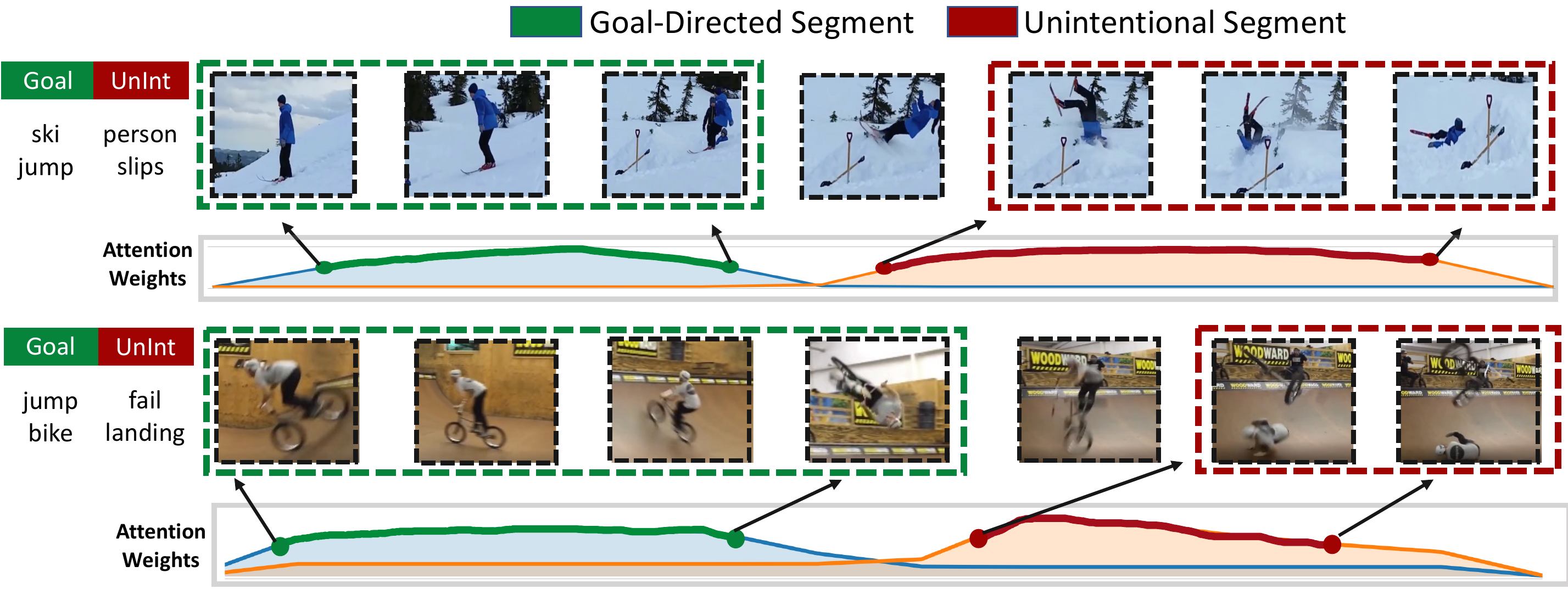} 
    \caption{\small Our method is able to identify the temporal regions that correspond to goal-directed/unintentional activity via the produced weighted TCAMs. Blue and Orange attention maps correspond to the goal-directed action and unintentional action respectively. }
    \label{fig:qual-analysis}
\end{figure*}

 \begin{figure}[t]
 \centering
    \includegraphics[width=0.45\textwidth]{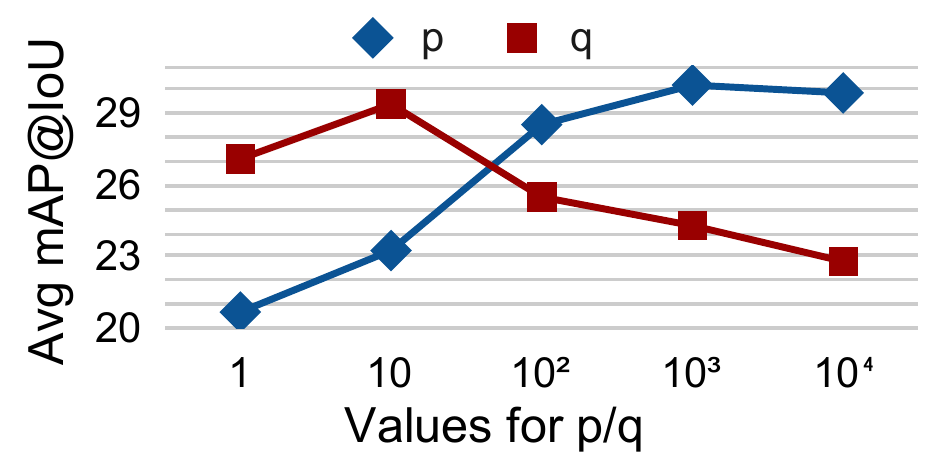} 
    \vspace{-2mm}
    \caption{\small Effect on Average (Goal+UnInt) mAP@IoU for the goal-directed and unintentional action when changing $p$ (blue) and $q$ (red). }
    % \jacob{this picture still takes too much space, put 4 curves into one figure or move this ablation in the appendix totally?}}
    \vspace{-4mm}
    \label{fig:p-ablation}
\end{figure}

% \begin{figure}[t]
%     \includegraphics[width=0.45\textwidth]{figure/ablation-q.pdf} 
%     \vspace{-3mm}
%     \caption{\small Effect on Average IoU for the goal-directed and unintentional action when changing $q$ }
%     \vspace{-4mm}
%     \label{fig:q-ablation}
% \end{figure}

\subsection{Goal-directed/Unintent. Action Localization}
Our model should be able to focus on the correct regions of the video in order to infer the goal-directed and unintentional action segments, hence understanding the transition between these two.
In order to evaluate our model on the task of localizing goal-directed as well as unintentional segments, we follow the standard evaluation protocol for temporal localization tasks by calculating the mean average precision (mAP) over different intersection over union (IoU) thresholds for both the types of actions. Since there are no quantitative results reported on our dataset, we use competitive models from the traditional weakly supervised action localization task as baselines. Since these models are trained using only one classification head which is used to identify the atomic actions in the video, we repurpose these models by adding an additional classification head (for the goal-directed and unintentional action) and bottom-up attention module (in the case of STPN~\cite{nguyen2018weakly}) or additional branch (in case of A2CL-PT~\cite{wsal2}) to adapt it to our task. We then retrain these models on our dataset and report quantitative results for comparison in Tab.~\ref{table:1}. It may be noted that our method performs significantly better than other weakly supervised methods on this task, when using the same backbone. For example, the average mAP@IoU score of our method outperforms A2CL-PT by 12.8\% for the goal-directed action and 13.5\% for the unintentional action, when using an I3D backbone. We conjecture that this localization improvement is due to our overlap regularization on the bottom-up attention weights since it enforces the model to focus on distinct portions of the action scene while ensuring the temporal order of the actions, which is a crucial property for solving this task. The qualitative results (see Appendix) show how the WSAL models focus on overlapping regions when inferring the goal-directed/unintentional action which reduces it's localization performance.
% \jacob{add more specific number comparisons here, say Our's reaches XX while STPN just XX. We conjecture that this improvement comes from our's XX design, which XXX. (Give more analysis here.)}

\subsection{Goal-directed/Unintent. Action Classification}
Given any video our model is trained to predict the goal-directed action as well as the unintentional action it eventually transitions into. Following previous works \cite{nguyen2018weakly, wsal6}, we use mean average precision (mAP) to evaluate the classification performance of our model on predicting the goal-directed action as well as unintentional action. We report our results in Tab.~\ref{table:2}. It is interesting to note that our method performs the best on the classification task as well. For example, it performs 4.1\% higher on the Goal cMAP and 6.3\% higher on the Unintentional cMAP than A2CL-PT when using an I3D backbone.

\subsection{Ablation Study} \label{ablation}
We conduct an ablation study to analyse various components of our model. We analyse the significance of the overlap regularization introduced in Section~\ref{overlap}. We observe very clearly in Tab.~\ref{table:4} that only using $\mathcal{L}_{cls}$ is not sufficient to localize the goal-directed and unintentional actions, and our final model performs the best. This implies that all components are necessary in order to achieve the best performance and each one is effective.
We further analyse the importance of the hyper-parameters $p$ and $q$ used in the overlap regularization in Fig.~\ref{fig:p-ablation}. We can see that increasing $p$ from 1 to $10^3$ results in a significant increase in the average mAP@IoU. This shows that the localization performance increases by penalizing the overlap of the bottom-up attentions more, but plateaus after the $10^3$ mark.
Analysing the $q$ hyper-parameter, we notice that increasing the value of $q$ decreases the performance. Since increasing the value of $q$ results in a lower margin of separation between the expectations of the goal-directed and unintentional bottom-up attention weights, we can conclude that a lower value of $q$, \ie, higher margins of separation helps achieve a better localization performance. However, $q=1$ signifies the extreme case when the margin is equal to the length of the clips, forcing the goal-directed and unintentional attention maps to be at two separate ends of the temporal axis, thereby hurting the performance. Fig.~\ref{fig:qual-analysis} shows qualitative examples of localizing the goal-directed and unintentional segments on our W-Oops dataset. We further provide more qualitative examples in the Appendix, which compare our method with previous WSAL methods.

\begin{figure}[t!]
    \centering
    \includegraphics[width=.45\textwidth]{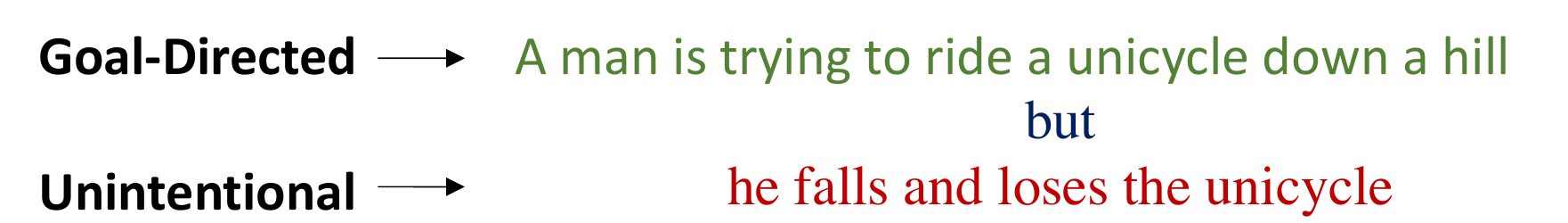} 
    \caption{\small We form the ground truth caption by concatenating the goal-directed and unintentional caption with a \textit{but}. }
    \label{fig:gt-cap}
\end{figure}

\subsection{Video Captioning}
\begin{figure}[t!]
    \centering
    \includegraphics[width=.45\textwidth]{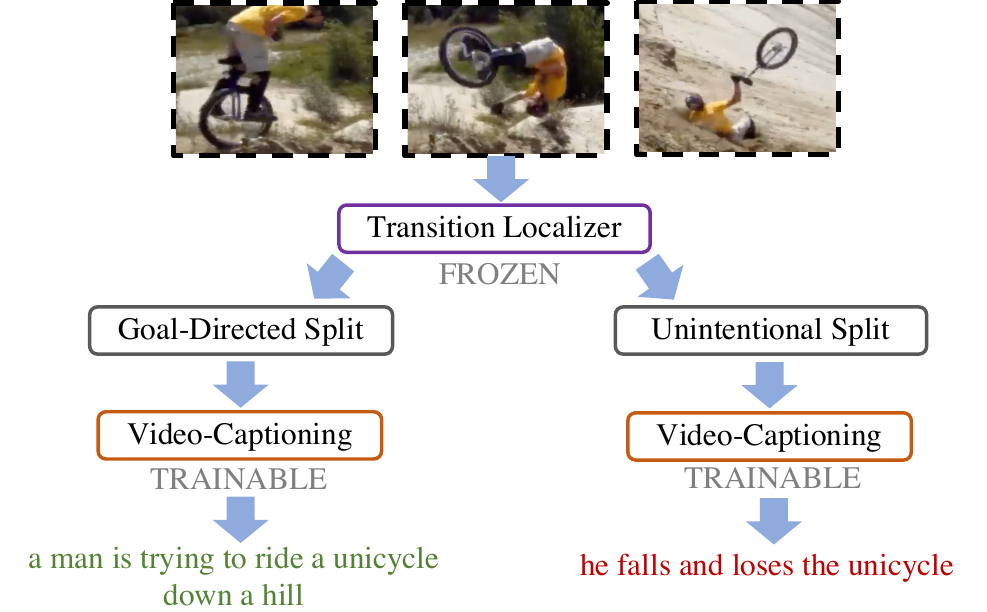} 
    \caption{\small Pipeline for the captioning experiment (using our localization module). 
    % \jacob{reduce the height of this picture, this pipeline figure is way too big. Only use half of its current space or move it to the appendix? }
    }
    \label{fig:cap-exp-pipeline}
\end{figure}

Teleological understanding helps explain the action better especially when the goal is partially/not achieved. In order to further verify whether our localization module can assist in teleological understanding, we conduct a captioning experiment which involves explaining the goal-directed and unintentional parts of the video through natural language descriptions.
% Hence, we conduct a video captioning to verify whether our localization module can assist in teleological understanding by explaining the goal-directed and unintentional parts of the video through natural language descriptions.
% further evaluate the teleological ability of our model by conducting a captioning task to verify whether the usage of our localization model can help explain the video better in terms of natural language descriptions.
% We further evaluate the teleological ability of our model by conducting a captioning task to verify whether the usage of our localization module can help improve captioning on videos containing unintentional actions.
We obtain ground truth captions for the goal-directed and unintentional regions from~\cite{Epstein_2020_CVPR}. The goal-directed and unintentional captions are concatenated (shown in Fig.~\ref{fig:gt-cap}) to form the ground truth caption for the entire video. We use an $80\%$-$20\%$ train/test split obtaining $3,200$ training samples and $800$ test samples. Our experimental setup is divided into two parts:
\begin{itemize}[leftmargin=*]
  \setlength{\itemsep}{0pt}
\item Train a state-of-the-art video captioner on the entire video and the whole ground truth caption.
\item Automatically split the video into the goal-directed and unintentional regions (inferred from their respective attention maps) using our trained localization module. Train two video captioners with distinct weights, to caption the goal-directed and unintentional regions with their respective captions as training signals. Finally, concatenate the two outputted captions by a \textit{but} to form the final predicted caption (during inference). Pipeline shown in Fig.~\ref{fig:cap-exp-pipeline}.
\end{itemize}

We use RMN~\cite{tan2020learning} as our captioning module, and extract features in the same manner as mentioned by the authors. Note that the localization module is kept frozen throughout the training process. 
From Tab.~\ref{table:5}, we observe that the results obtained after splitting the video into its goal-directed and unintentional region outperform those when trained on the entire video. In particular, our method with transition localizer significantly outperforms the baseline without localizer by $5.1$ SMURF~\cite{feinglass2021smurf} scores and $4.1$ CIDEr~\cite{vedantam2015cider} scores.
Notably, SMURF~\cite{feinglass2021smurf} is a caption evaluation algorithm that incorporates diction quality into its evaluation, which demonstrates SOTA correlation with human judgment and improved explainability. Through SMURF, we observe that our method improves semantic performance while maintaining the descriptiveness of terms used in the sentence. %This leads to a large improvements in caption quality that are not adequately captured by purely semantic metrics.
These metrics clearly show that the teleological ability of our localization module helps a captioning module output more accurate captions on videos containing unintentional actions, which is achieved by the precise capture of unintentional activity in video. 
% We showcase more qualitative examples in the Appendix.

\begin{table}[t!]
\centering
\setlength{\tabcolsep}{6pt} % Default value: 6pt
{\small
 \begin{tabular}{c|c c c c} 
% \toprule \\ [-3.8ex]
 \textbf{Exp.} & R~\cite{lin2004rouge} & M~\cite{banerjee2005meteor} & C~\cite{vedantam2015cider} & S~\cite{feinglass2021smurf} \\ 
 \hline
 \hline
 Without Loc. & 16.7 & 37.7 & 29.8 & 14.1 \\ 
%  \hline
 With Loc. & \textbf{17.0} & \textbf{38.0} & \textbf{33.9} & \textbf{19.8} \\
%  \bottomrule
 \end{tabular}
 \vspace{-1mm}
 \caption{\small Captioning results with or without localization module. R, M, C, S denote ROUGE\_L, METEOR, CIDEr and SMURF metrics respectively for video captioning evaluations.}
 \label{table:5}
   \vspace{-4mm}
 }
\end{table}

 \section{Conclusion}
In this paper, we propose W-Oops, an augmented unintentional human  activity dataset that consists of both goal-directed and unintentional video-level activity annotations, built upon Oops \cite{Epstein_2020_CVPR}. We consider a weakly supervised task to infer the respective classes as well as the temporal regions in which they occur using only the video-level activity annotations. We further build a neural network architecture which employs a novel overlap regularization on top of the bottom-up attention weights outputted by our attention module, which helps the model focus on distinct parts of the video while maintaining the temporal ordering of these actions when inferring the temporal regions. We conclude from our experiments that out method significantly outperform previous WSAL baselines on our benchmark. The video captioning experiment further verifies the teleological ability of our localization module, which points a promising future research direction for improving captioning quality through teleological analysis. 

{\small
\noindent \textbf{Acknowledgement.}
This work was supported by the National Science Foundation under Grant CMMI-1925403, IIS-2132724 and IIS-1750082. 
}

\def\abstract
 {%
 % Suppress page numbers when the appropriate option is given
 \iftoggle{cvprpagenumbers}{}{%
\thispagestyle{empty}
 }
 \centerline{\large\bf Supplementary Materials}%
\vspace*{12pt}%
 \it%
 }

\def\endabstract
 {
 % additional empty line at the end of the abstract
 \vspace*{12pt}
 }

\vspace{8mm}
\begin{abstract}
This document provides additional details and further analysis of our model architecture. We start by providing detailed statistics about the W-Oops dataset in Sec.~\ref{dataset-stats}. We further analyse the dependence of unintentional actions on goal-directed actions in Sec~\ref{entropy-sec}. We then give more details on the choice of our backbone features and experimenting with human-pose features in Sec.~\ref{pose-sec} . We further analyse the video embedding module by removing it entirely/replacing it with a Transformer Encoder~\cite{NIPS2017_3f5ee243} in Sec.~\ref{trans-sec}. We study the effect of different selections of $\lambda$, the hyperparameter that controls the trade-off between our losses in Sec.~\ref{lambda-sec}. Details into the 3D-CNN feature extraction is provided in Sec.~\ref{3d-sec}. Finally, we explain more about our annotation tool in Sec.~\ref{annottool-sec}, and provide additional qualitative results for the localization and captioning experiments in Sec.~\ref{qual-sec} and Sec.~\ref{cap-qual-sec}.
\end{abstract}
% respectively.

\section{W-Oops Statistics and Analysis}\label{dataset-stats}
\begin{figure*}[t!]
\centering
    \includegraphics[width=\linewidth]{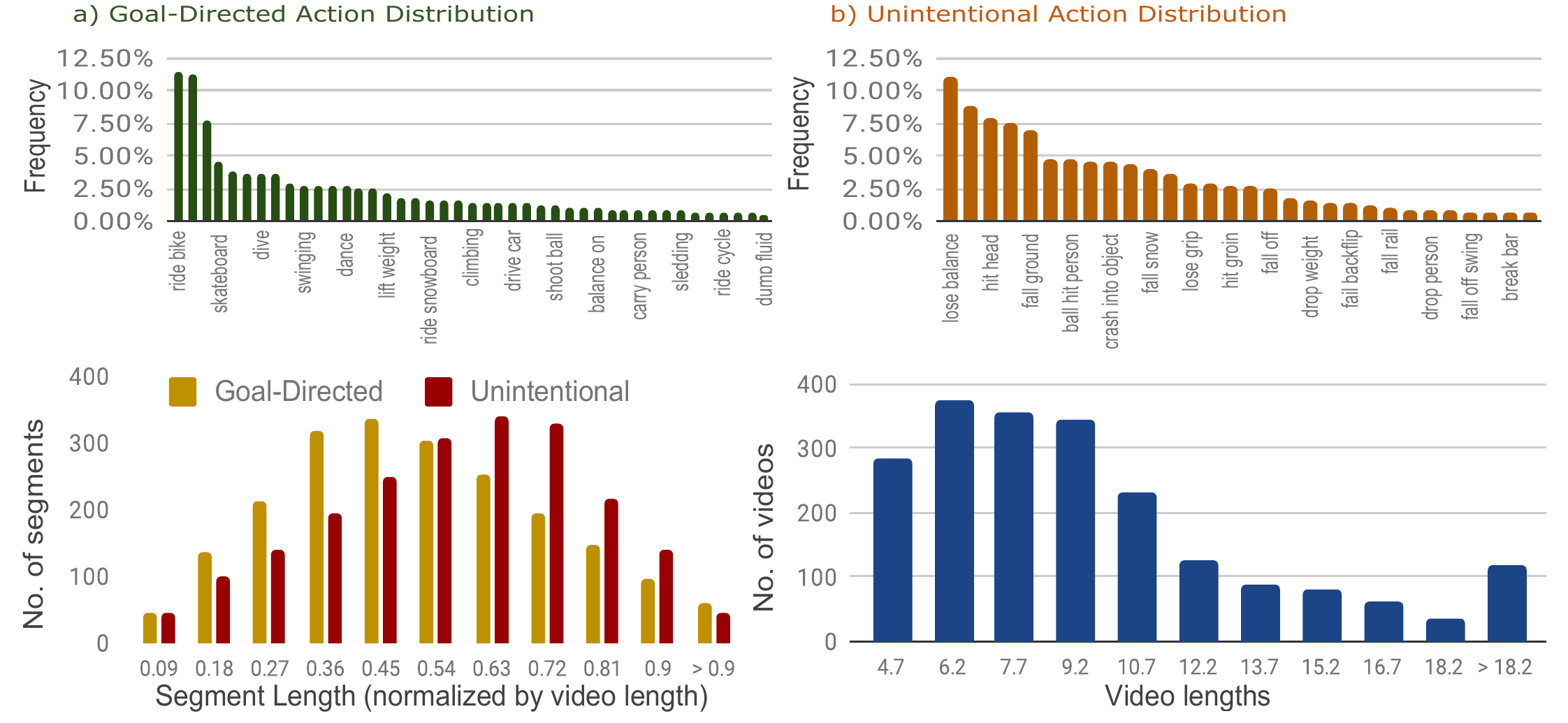} 
    % \

    % \caption{\small Distribution over the goal-directed and unintentional actions. All action classes are not shown due to space constraints. Please refer to the appendix in order to view the full class list.}
    % \label{fig:class-dist}
    % \vspace{1mm}
    \caption{\small Top: Distribution over the goal-directed and unintentional actions (partially shown). Lower Left: Distribution over goal-directed and unintentional segment lengths (normalized by the video length). Lower Right: Distribution over the entire video length. }
    % \
    \label{fig:len-dist}
\end{figure*}

The final W-oops dataset contains 1582 train samples and 526 testing samples, containing a total of 44 diverse goal-directed and 30 unintentional action classes, as seen in Fig.~\ref{fig:len-dist}. We have also provided the distribution of the goal-directed and unintentional segment lengths, as well as the total video lengths. It shows that the goal-directed and unintentional segment lengths are well diversified over then entire length of the video. The lengths of the video are short in general, with a majority of them ranging from 6.2 - 7.7 seconds. This makes the task of identifying these sub-regions in the video challenging.
In our benchmark, train samples contain only video-level labels whereas the test samples contain both the video-level labels as well as the unintended activity transition points (taken from the original Oops dataset), which we use to split the video into a goal-directed and unintentional region in order, for evaluation.
% to conduct evaluation.

\section{Can Unintentional Actions be predicted knowing the Goal-Directed Action?}\label{entropy-sec}
\begin{figure*}[t!]
\centering
    \includegraphics[width=0.95\linewidth]{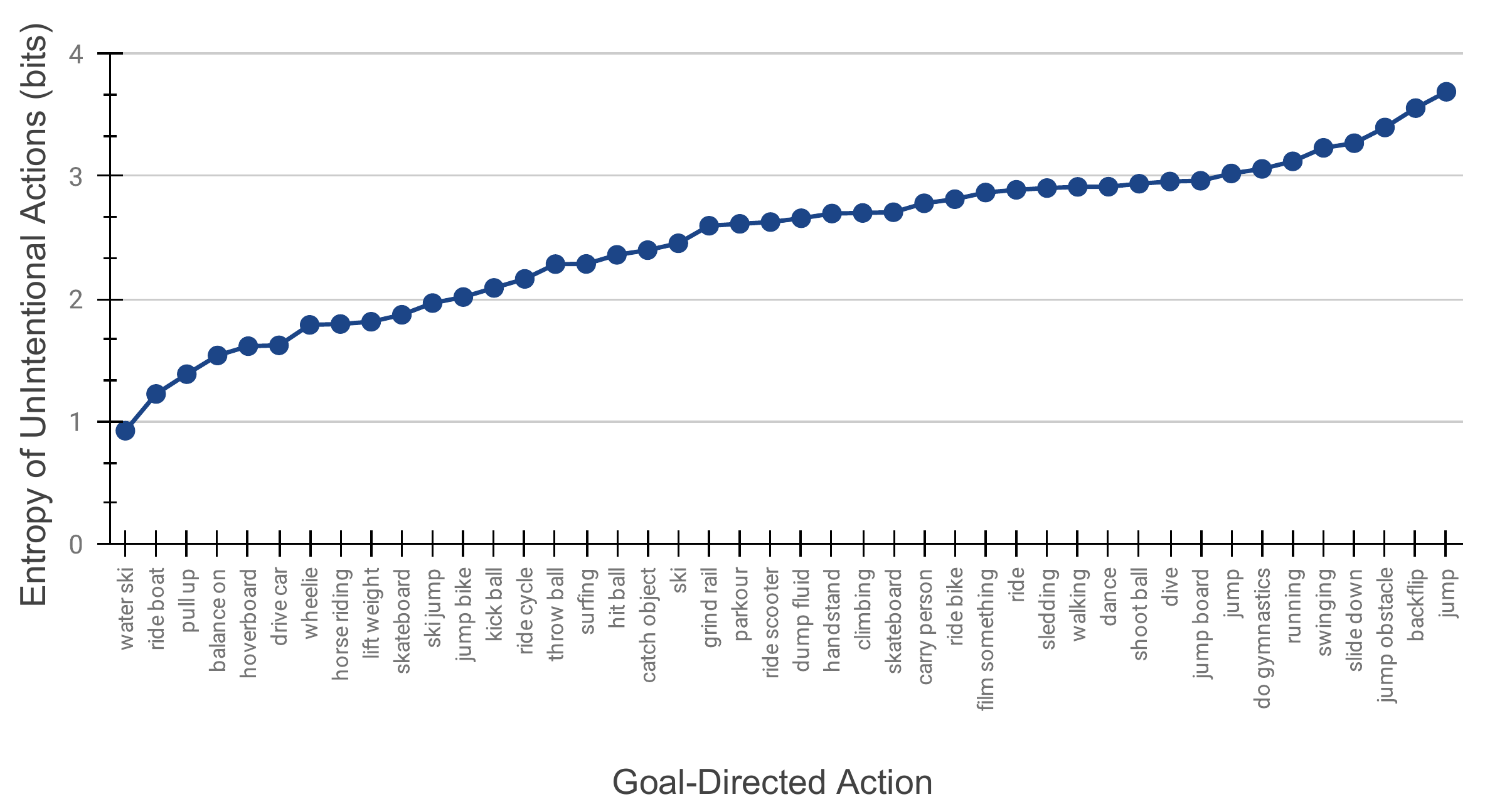} 
    % \
    \caption{\small Entropy (in bits) of the unintentional actions conditioned on the goal-directed actions. We can see that the unintentional actions are correlated to the goal-directed actions but are not completely predictable. }
    % \
    \label{fig:entropy_uinint}
\end{figure*}

In this section we analyse the amount of information knowing about a goal-directed action gives us when inferring the unintentional action. In order to do this, we calculate a probability distribution of the unintentional actions conditioned on the goal-directed actions and calculate their entropy. An entropy of 0 would indicate that the unintentional action can predicted from the goal-directed action alone. On the other hand, an entropy of 4.91($-\log_{2}(30)$) indicates that the unintentional actions are uncorrelated with the goal-directed action. Fig.~\ref{fig:entropy_uinint} shows us that the conditional entropy of unintentional actions lies between these two values, suggesting that they are correlated but are not completely predictable knowing the goal-directed action.

\section{Using 2D Pose Features}\label{pose-sec}
Successful attempts at using human skeleton features for activity recognition.\cite{pose1,pose2,pose3}, fall prediction \cite{fall1, fall2} and action localization \cite{skeleton-loc1} provides encouragement to use them for our task as well. However human skeleton features alone would not be enough as it does not capture the surrounding environment information which the RGB features do. Hence we concatenate both the RGB features and skeleton features to use as our backbone features.\par
\begin{figure*}[t!]
\centering
    \includegraphics[width=0.95\linewidth]{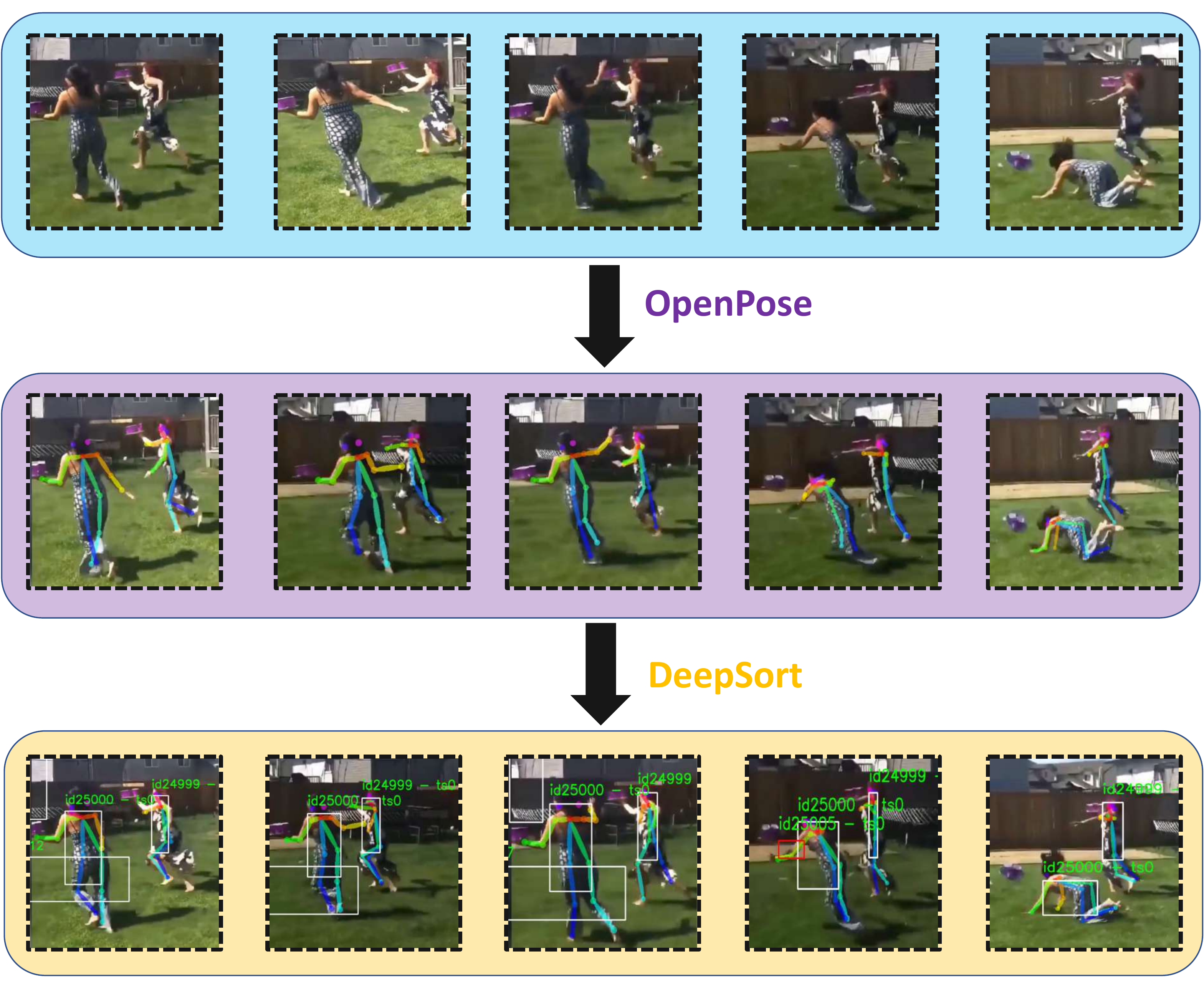} 
    % \
    \caption{\small An example of extracting body keypoint coordinates of multiple agents in videos using Openpose~\cite{cao2019openpose}, followed by Deepsort~\cite{wojke2017simple} to cluster the keypoints of the same person across the frames.}
    % \
    \label{fig:openpose}
\end{figure*}

In order to test this hypothesis, for each video we extract 2D keypoint coordinates of human(s) from each observed frame using OpenPose \cite{cao2019openpose}. Since OpenPose is able to capture multiple human(s) in a frame, we use DeepSort \cite{wojke2017simple} to cluster the keypoints of the same person across frames. We denote the sequence of observed keypoints from the $i^{th}$ person in the video as $\mathbf{K^{i}} = (\mathbf{k}_1^{i}, \mathbf{k}_2^{i}, . . ., \mathbf{k}_t^{i})$, where $\mathbf{k}^i_j$ denotes the keypoint coordinates of the $i^{th}$ person in frame $j$, with $t$ being the total number of frames. Example showed in~\ref{fig:openpose}\par 
Using the COCO model of OpenPose, we obtain 18 keypoint coordinates for each observed person in a frame, which include coordinates for the nose, neck, left and right shoulders, hips, elbows, wrists, knees, ankles, eyes and ears, i.e., each 
\begin{equation}
    \mathbf{k}^i_j = (x^i_{j,1},y^i_{j,1},x^i_{j,2},y^i_{j,2}, . . . , x^i_{j,18},y^i_{j,18})
\end{equation}
Since these coordinates do not capture the correlation between different keypoints, we follow the process in \cite{fall2} to vectorize these coordinates to incorporate these correlations. We ignore the face keypoints (eyes, ears and nose), since we want to focus only on the body pose. We then transform the remaining 13 coordinates into vectors connecting the adjacent keypoints as illustrated in Fig. The shoulders are connected to the neck, elbows are connected to the corresponding shoulders, wrists are connected to corresponding elbows, hips to the neck, knees to the corresponding hips and finally the ankles to the corresponding knees. Following this process as followed in \cite{fall2}, we obtain 12 keypoint vectors from the 13 keypoint coordinates, and normalize them to unit length. For the $m^{th}$ connection pointing from the $p^{th}$ keypoint to the $q^{th}$ keypoint, the keypoint vector $(\overline{x^i_{j,m}},\overline{y^i_{j,m}})$ for the $i^{th}$ person in frame $j$ is calculated as:
\begin{equation}
    (\overline{x^i_{j,m}},\overline{y^i_{j,m}}) = \dfrac{(x^i_{j,q}-x^i_{j,p},y^i_{j,q}-y^i_{j,p})}{\sqrt{{(x^i_{j,q}-x^i_{j,p})}^2 + {(y^i_{j,q}-y^i_{j,p})}^2}}
\end{equation}
We calculate this for each of the 12 connections, and concatenate them to get:
\begin{equation}
    \overline{\mathbf{k}^i_j} = (\overline{x^i_{j,1}},\overline{y^i_{j,1}},\overline{x^i_{j,2}},\overline{y^i_{j,2}}, . . . , \overline{x^i_{j,12}},\overline{y^i_{j,12}})
\end{equation}

Videos involving action such as two people colliding with another person, or a person carrying another person, requires features of multiple people in order to understand these actions. Hence we concatenate the keypoints of the two most frequently occurring people $l$ and $r$ as detected by DeepSort, and concatenate them to get the final feature vector for frame $j$ as $\overline{\mathbf{k}_j} = \overline{\mathbf{k}^l_j} \oplus \overline{\mathbf{k}^r_j}$. \par
Note, that there may be partially missing or completely missing keypoint coordinates for a person in a certain frame. In the case of partially missing keypoints we set a keypoint vector containing a connection to a missing keypoint to (0,0). In the case of completely missing keypoints we set all the keypoint vectors to (0,0) in the case the person had not been detected yet, or else set all the keypoint vectors to the corresponding last observed keypoint vectors of the person. \par
RGB features are extracted by passing non-overlapping chunks of 16 frames to a pretrained 3D CNN architecture. Since the skeleton feature are extracted for each frame, we concatenate  skeleton features extracted from consecutive and non-overlapping chunks of 16 frames. We convert $\overline{\mathbf{k}} = (\overline{\mathbf{k}_1},\overline{\mathbf{k}_2}, . . ., \overline{\mathbf{k}_t})$ to $\widetilde{\mathbf{k}} = (\widetilde{\mathbf{k}_1},\widetilde{\mathbf{k}_2}, . . ., \widetilde{\mathbf{k}_{t/16}})$, where $\widetilde{\mathbf{k}_h}$ for the $h^{th}$ chunk is given by :
\begin{equation}
    \widetilde{\mathbf{k}_h} = \overline{\mathbf{k}_{16(h-1)+1}} \oplus \overline{\mathbf{k}_{16(h-1)+2}} \oplus . . . \oplus \overline{\mathbf{k}_{16(h)}}
\end{equation}
\par
We finally concatenate the RGB features $X$ and the skeleton features $\widetilde{\mathbf{k}}$ to obtain $X_{cat} = (X_1 \oplus \widetilde{\mathbf{k}_1}, X_2 \oplus \widetilde{\mathbf{k}_2}), . . . , X_l \oplus \widetilde{\mathbf{k}_l})$, where $l$ is the total number of 16 frame chunks (clips) in the video.

We then provide comparisons between using only the RGB features and using the RGB features concatenated with the skeleton features in table \ref{table:1} . We can see that the performance decreases, from 35.0\% to 34.7\% for the goal-directed mAP@IoU and from 26.0\% to 24.6\% for the unintentional mAP@IoU. We conjecture that this performance decrease is due to the noise introduced by the incorrect/missing keypoint coordinates at certain frames, as well as due to some of the videos which involve an agent driving a vehicle and hence the agent is partially or completely not seen in the video

\begin{table}[h!]
\centering
\setlength{\tabcolsep}{3.8 pt} % Default value: 6pt
\renewcommand{\arraystretch}{1.2} % Default value: 1
{\small 
 \begin{tabular}{c|c| c c c c} 
%  \toprule \\ [-3.8ex]
\multirow{2}{*}{Feature} & \multirow{2}{*}{Segment} & \multicolumn{4}{c}{\textbf{mAP@IoU}} \\ 
& & 0.3 & 0.5 & 0.9 & Avg \\[0.2ex] 
 \hline
\multirow{2}{*}{RGB (I3D)} & \text{Goal} & 49.9 & 41.1 & 5.0 & 35.0\\
& \text{UnInt} & 36.4 & 30.0 & 2.8 & 26.0\\
\multirow{2}{*}{RGB (I3D) + Skeleton} & \text{Goal} & 47.1 & 42.1 & 4.8 & 34.7\\
& \text{UnInt} & 34.4 & 27.4 & 2.1 & 24.6\\
%  \bottomrule
 \end{tabular}
%  \vspace{-1mm}
 \caption{\small Analysis of the effect of skeleton features.}
 \label{table:1}
 }
\end{table}

% \begin{abstract}
%      In videos that contain actions performed unintentionally, agent do not achieve their desired goals. It is therefore challenging for computer vision systems to understand high-level concepts such as goal-directed behavior from such videos. On the other hand, humans from a very early age, are able to understand the relation between an agent and their ultimate goals even if the action is disrupted or unintentional effects occur. To validate this ability of deep learning models to perform this task, we curate W-Oops, built on top of the original Oops dataset, which is a novel dataset consisting  of $\sim$ 2100 unintentional action videos containing 44 goal-directed as well as the 33 unintentional activity classes. This paper considers a weakly supervised task for localizing the goal-directed as well as unintentional part of a video using only these action labels. We also propose a novel neural network architecture to solve the same. 
      
% \end{abstract}
% \section{Test}

\section{Analysis of Video Embedding Module}\label{trans-sec}
We now analyse the effectiveness of our video embedding module, by removing the module and using only the raw features from the frozen feature extractor. We also compare our video embedding module which consists of a GRU with a Transformer Encoder \cite{NIPS2017_3f5ee243}, a component of the original Transformer architecture which has achieved state of the art results on many vision \cite{transformer-cv1,transformer-cv2,transformer-cv3,transformer-cv4,transformer-cv5,transformer-cv6} as well as NLP \cite{transformer-nlp1, transformer-nlp2, transformer-nlp3, transformer-nlp4} tasks. As opposed to a GRU which learns feature representations at each time step in a sequential manner by using the hidden state in the previous timestep, a transformer encoder uses multiheaded self attention to calculate the dependency of each token in the sequence to encode the token at the current timestep.
As seen in table \ref{table:2}, we can see that using static backbone features result in a very poor localization performance. Additionally it is also interesting to observe that the GRU performs better than the transformer.

\begin{table}[h!]
\centering
\setlength{\tabcolsep}{3.8pt} % Default value: 6pt
\renewcommand{\arraystretch}{1.2} % Default value: 1
{\small 
 \begin{tabular}{c|c| c c c c} 
%  \toprule \\ [-3.8ex]
\multirow{2}{*}{Embedding Module} & \multirow{2}{*}{Segment} & \multicolumn{4}{c}{\textbf{mAP@IoU}} \\ 
& & 0.3 & 0.5 & 0.9 & Avg(0.1:0.9) \\[0.2ex] 
 \hline
\multirow{2}{*}{None} & \text{Goal} & 30.2 & 16.5 & 1.3 & 18.7\\
& \text{UnInt} & 18.6 & 9.4 & 0.02 & 11.1\\
\multirow{2}{*}{Transformer Encoder} & \text{Goal} & 49.1 & 41.5 & 2.7 & 34.9\\
& \text{UnInt} & 31.7 & 17.9 & 0.7 & 22.7\\
\multirow{2}{*}{GRU} & \text{Goal} & 49.9 & 41.1 & 5.0 & 35.0 \\
& \text{UnInt} & 36.4 & 30.0 & 2.8 & 26.0 \\
%  \bottomrule
 \end{tabular}
%  \vspace{-1mm}
 \caption{\small Ablation study of the contribution of the video embedding module.}
 \label{table:2}
 }
\end{table}

\section{Analysis of Weight Tradeoff Parameter $\lambda$}\label{lambda-sec}
\begin{figure}[t!]
\centering
    \includegraphics[width=\linewidth]{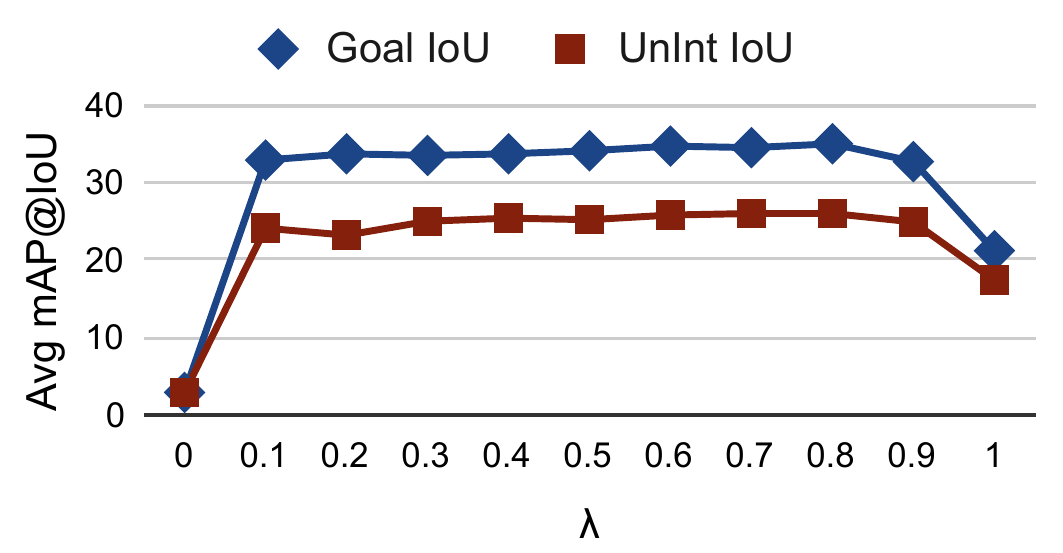} 
    % \
    \caption{\small Analysis of the effect of $lambda$ which is a scalar parameter for controlling the tradeoff between the MIL Loss and Overlap Regularization.}
    % \
    \label{fig:lambda-ablation}
\end{figure}

$\lambda$ is the scalar parameter used to control the tradeoff between the Multiple Instance Learning Loss (MIL) and the Overlap Regularization.
We study the effects of changing this parameter in the range of [0,1], where $\lambda$=0 corresponds to purely MIL Loss and $\lambda$=1 corresponds to purely Overlap Regularization. As seen in Fig.~\ref{fig:lambda-ablation}, we notice that for $0.3\leq\lambda\leq0.8$, the average mAP@IoU for the goal-directed and unintentional action remains almost constant, but on close observation we see that $\lambda = 0.8$ performs the best for the goal-directed as well as unintentional action.

\section{Feature Extraction Details}\label{3d-sec}
This section provides detailed explanation about the feature extraction process. We follow previous work~\cite{Epstein_2020_CVPR} and down-sample all raw videos at 25 FPS. We then create chunks of 16 consecutive and non-overlapping frames. In order the extract the I3D and R(2+1)D features, we pass these chunks to the respective backbone networks and obtain the features as the output of their global pooling layers.
We use the following libraries to extract R(2+1)D\footnote{\url{https://pytorch.org/vision/0.8/models.html}} and I3D\footnote{\url{https://github.com/deepmind/kinetics-i3d}} features from the videos.\par
\noindent\textbf{I3D}: For the I3D~\cite{carreira2017quo} features, we re-scale all frame pixels between -1 and 1, after which we resize the frames preserving aspect ratio such that the smallest dimension is 256 pixels. We then apply center crop to obtain $224\times224$ frames. Chunks of 16 non-overlapping frames are then passed through the RGB stream of a I3D~\cite{carreira2017quo} backbone pretrained on the Kinetics dataset~\cite{kay2017kinetics} to obtain features $\mathbf{X_i} \in \real^{1024\times l_i}$ from the global pooling layer.\par

\noindent\textbf{R(2+1)D}: For the R(2+1)D~\cite{tran2018closer} network, we re-scale frame pixels between 0 and 1, after which we resize all frames to $128\times171$. We then normalize these frames and finally apply center crop to obtain $112\times112$ frames. We the chunk the frames in the same way and pass it through the R(2+1)D~\cite{tran2018closer} backbone pretrained on Kinetics to obtain features $\mathbf{X_i} \in \real^{512\times l_i}$ from the global pooling layer.
\section{Annotation Tool for Label Evaluation and Correction}\label{annottool-sec}
\begin{figure}[t!]
\centering
    \includegraphics[width=\linewidth]{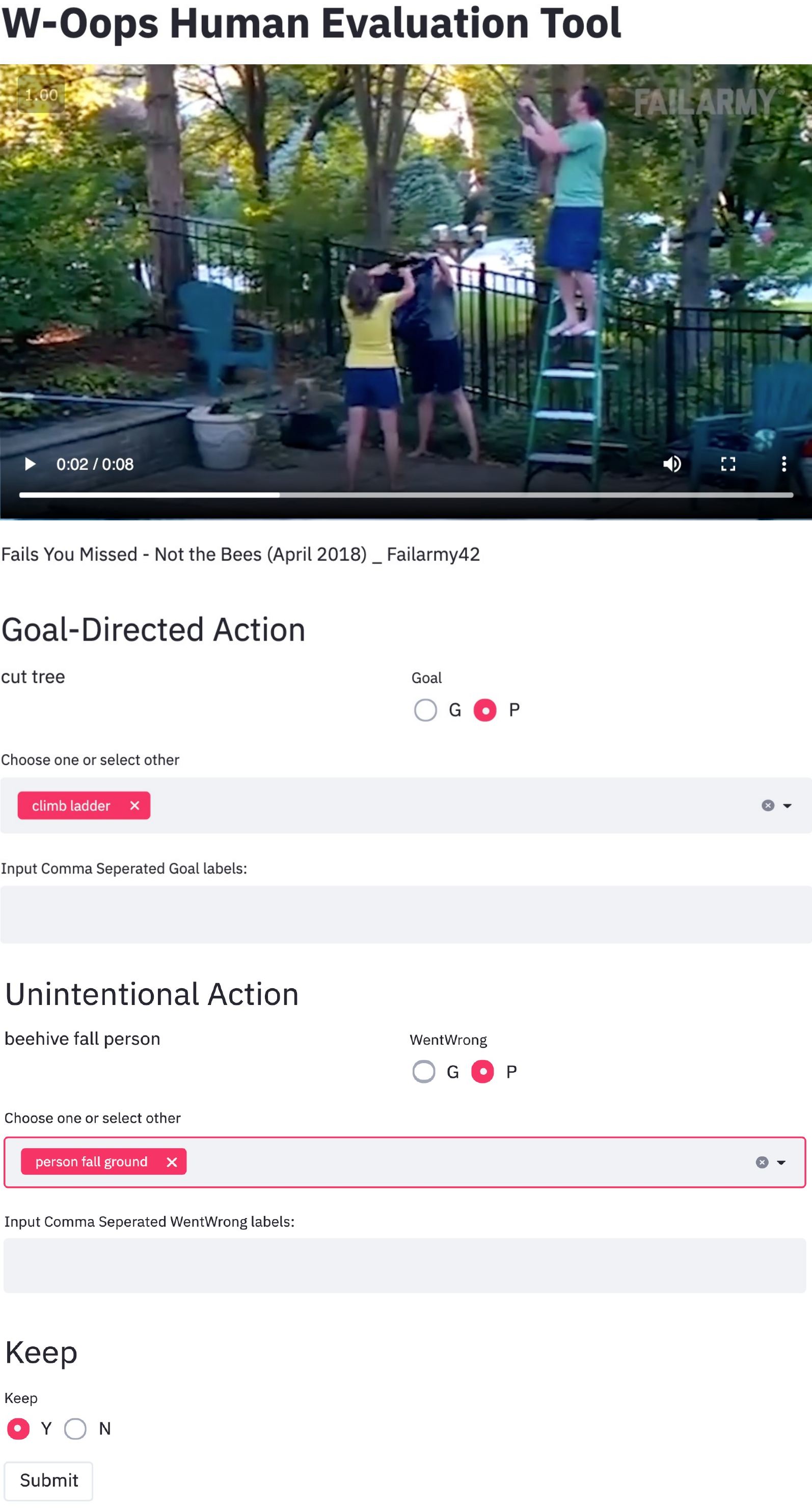} 
    % \vspace{1mm}
    \caption{\small Interface for W-Oops annotations, where we ask the annotators to rate the semi-automatically extracted goal-directed and unintentional actions as `Good' or `Poor'. If `Poor', they can choose from a fixed list of already present actions or input their own. They also have an option to indicate whether or not to keep the video in the case the goal in the video is ambiguous.}
    % \
    \label{fig:annotation-tool}
\end{figure}

The annotation tool used for the human evaluation and correction process is shown in Fig.\ref{fig:annotation-tool}. We provide a video to the evaluator along with the actions extracted from the annotations. The evaluator can then view the videos and mark the goal-directed actions as well as unintentional action as either `Good' (G) or `Poor' (P), with reference to the video. `Good' is given to an action which is entailed in the video and `Poor' otherwise. In case the evaluator marks an action as `Poor', they can then choose another action from the already present list of total actions, or else add a new action if not contained in the list. The evaluator also has an option to not keep the video in the case the goal of the agent in the video was ambiguous. 
Once this process is complete, evaluators can hit 'Submit', which would then load the next video.

\section{Qualitative Results of Goal-directed and Unintent. Action Localization}\label{qual-sec}
\begin{figure*}[t!]
\centering
    \includegraphics[page=1,width=0.95\linewidth]{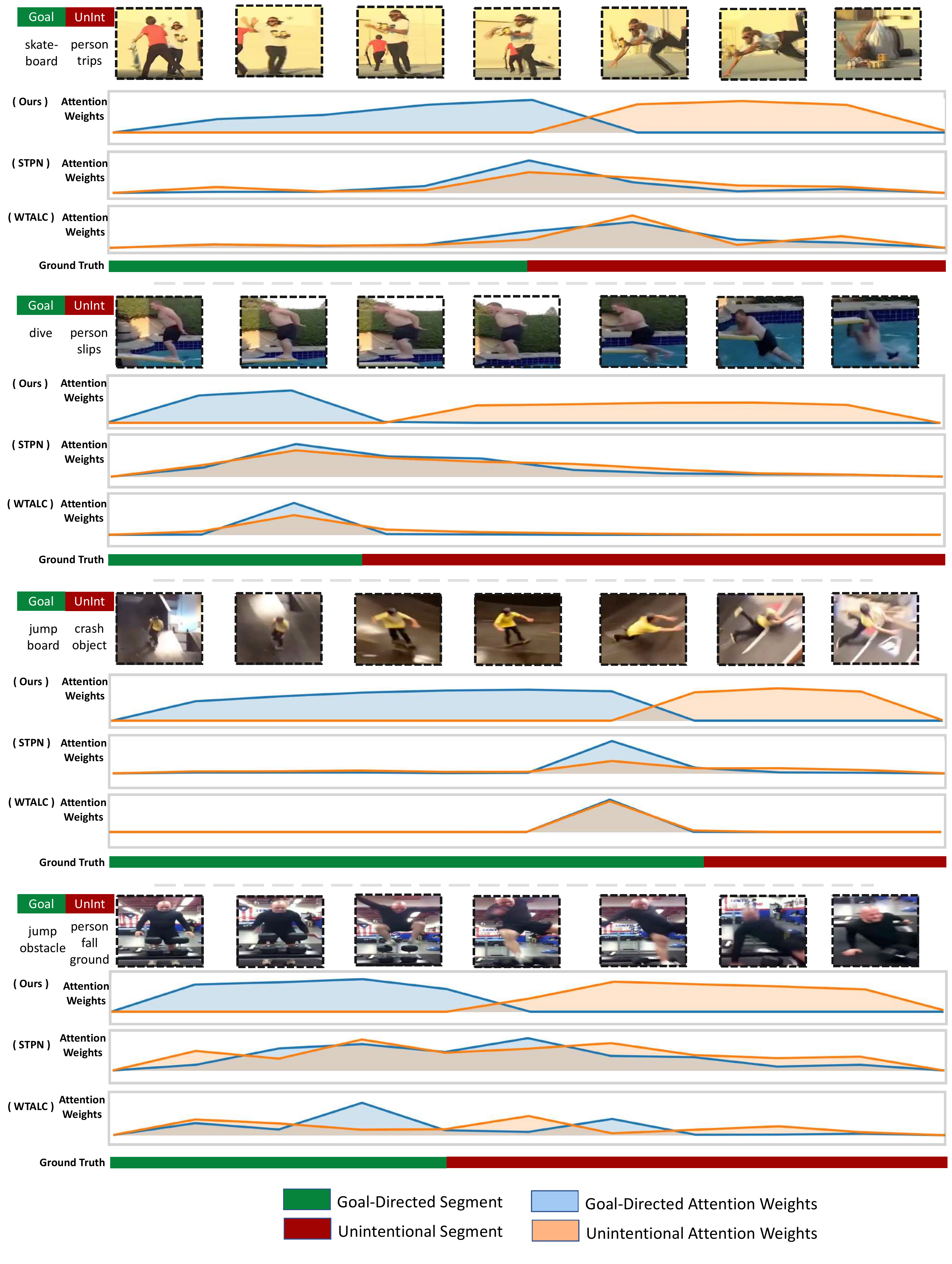} 
    
    % % \
    \caption{\small Qualitative results of our model's outputs. We provide attention weights outputted from STPN trained on our dataset, as well as the ground truth segments for comparison. }
    % \
    \label{fig:qual1}
\end{figure*}

\begin{figure*}[t!]
\centering
    \includegraphics[page=2,width=0.95\linewidth]{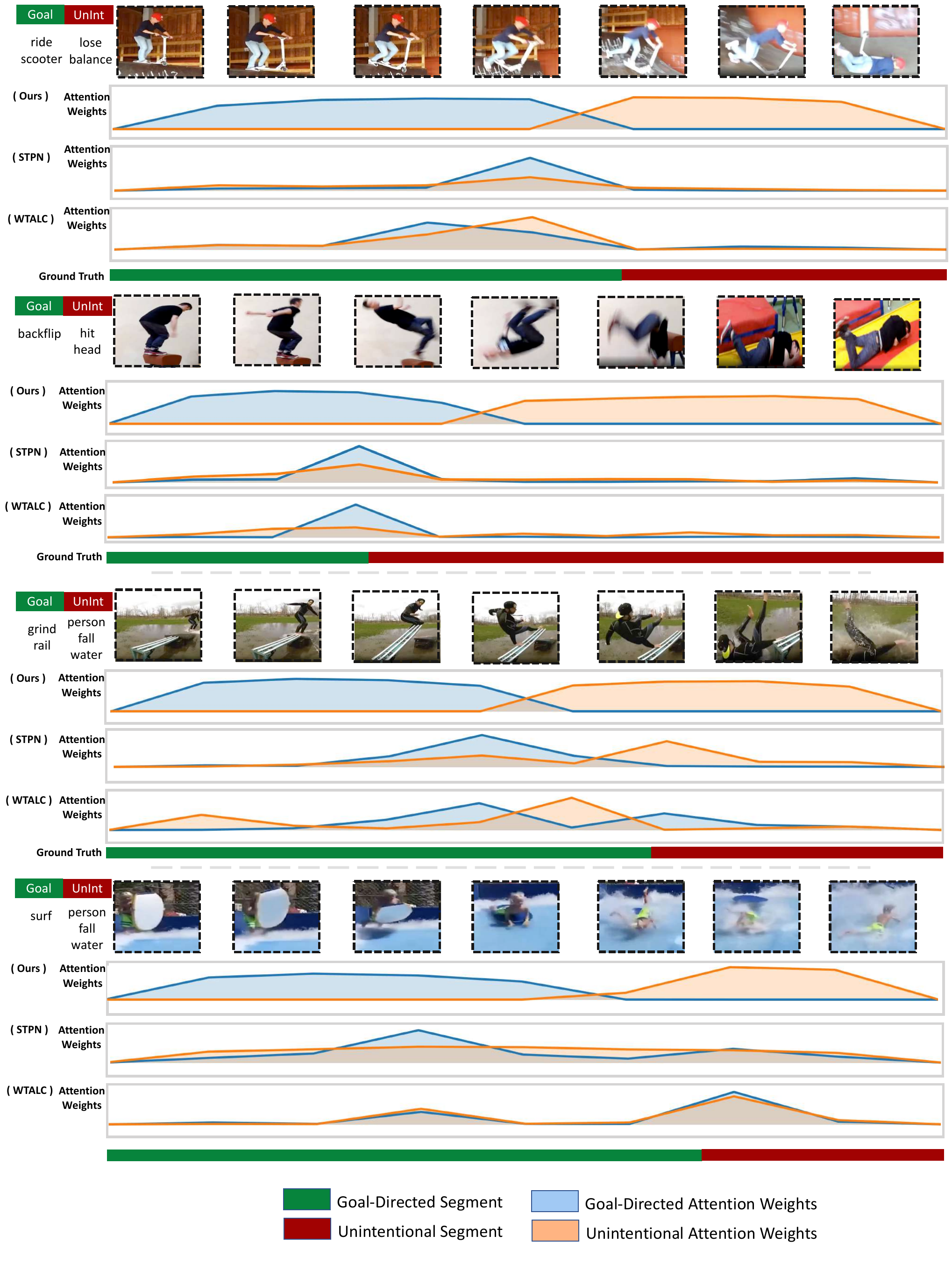} 
    % \
    \caption{\small Qualitative results of our model's outputs. We provide attention weights outputted from STPN trained on our dataset, as well as the ground truth segments for comparison. }
    % \
    \label{fig:qual2}
\end{figure*}

\begin{figure*}[t!]
\centering
    \includegraphics[page=1,width=\linewidth]{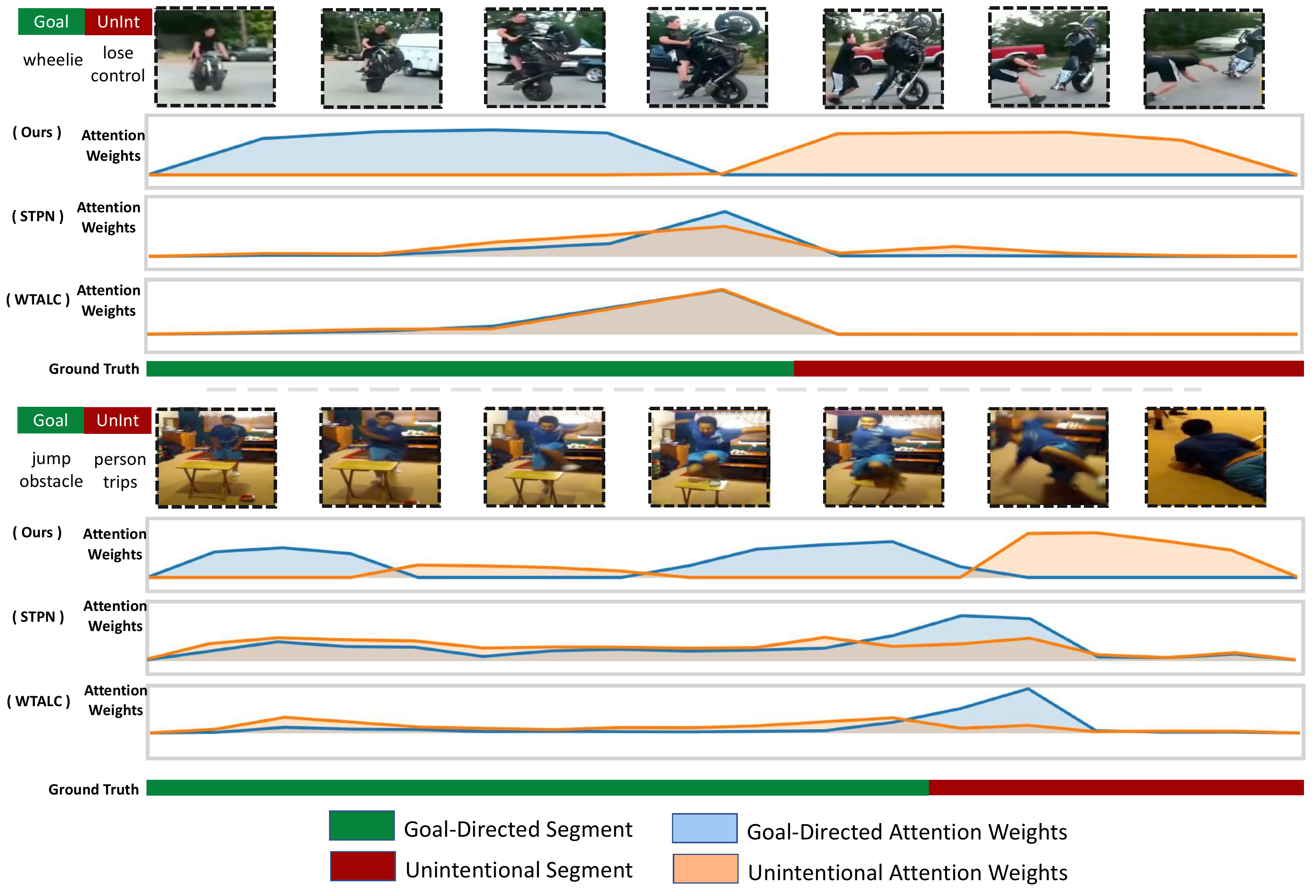} 
    % \
    \caption{\small Qualitative results of our model's outputs. We provide attention weights outputted from STPN trained on our dataset, as well as the ground truth segments for comparison. }
    % \
    \label{fig:qual3}
\end{figure*}

In this section, we provide additional qualitative results of our model, along with previous weakly supervised action localization (WSAL) models, namley WTALC~\cite{wsal6} and STPN~\cite{nguyen2018weakly}. We have provided examples of videos containing diverse actions, in order to show our model's generalizability. From Fig.~\ref{fig:qual1}, Fig.~\ref{fig:qual2} and Fig.~\ref{fig:qual3}, we notice that our model is able to focus on distinct regions in order to infer the goal-directed and unintentional actions, whereas the previous WSAL models focus on overlapping regions, and in many cases have very sparse attention weights. We conjecture this is due to the nature of task these models were originally built for, \ie, segmenting atomic actions from untrimmed videos. Additionally, we can see that the Overlap Regularization is able to enforce our model to maintain the temporal ordering of the goal-directed/unintentional action.

% \clearpage
\section{Qualitative Results for Video Captioning}\label{cap-qual-sec}
\begin{figure*}[t!]
\centering
\includegraphics[page=1,width=0.95\textwidth]{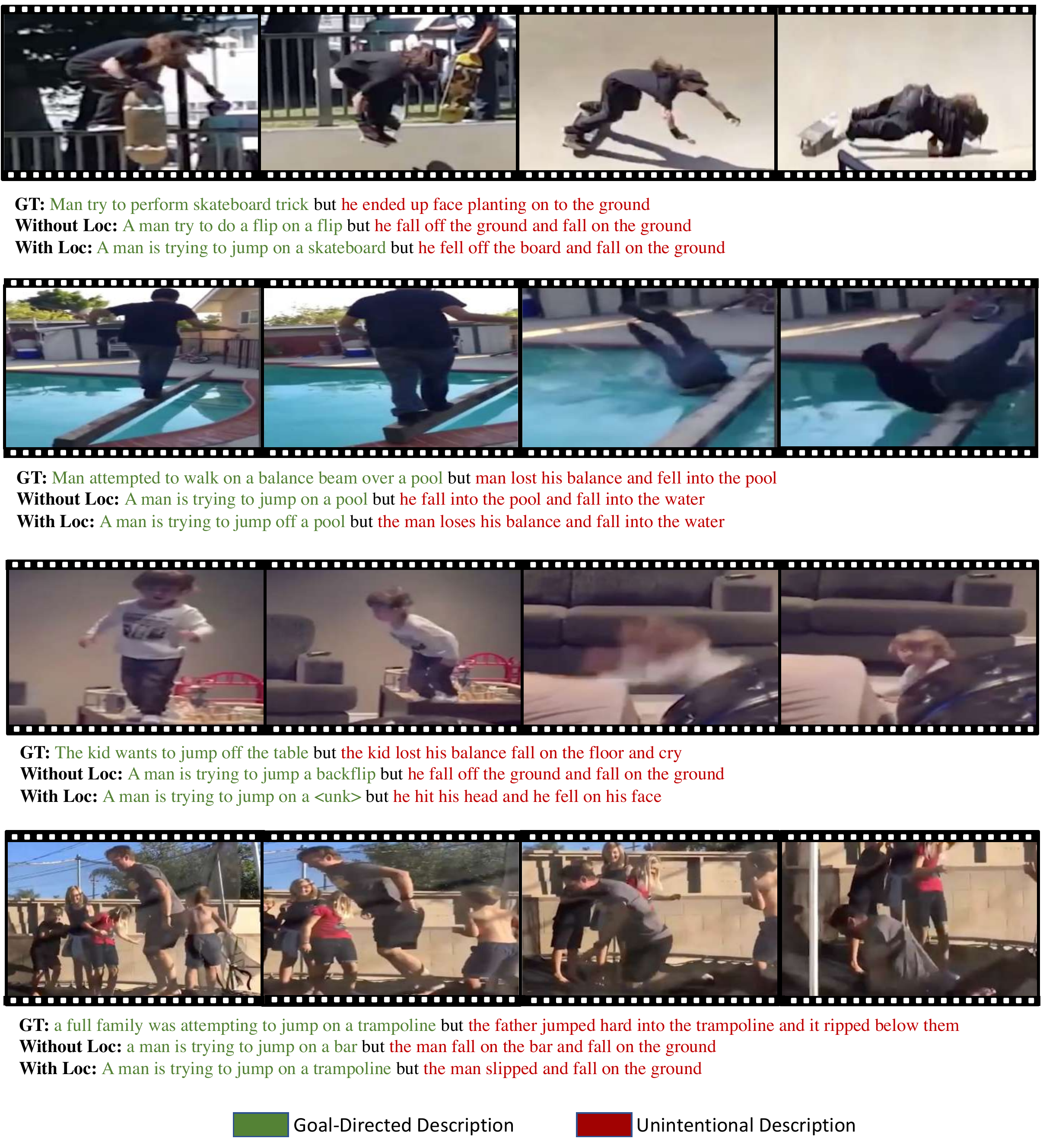} 
% % \
\caption{\small Qualitative results for the video captioning experiment. We provide ground truth captions from a human annotator, captions generated without as well as with our localization module. We observe that the captions generated leveraging our localization module tend to be more descriptive and semantically correct. }
% \
\label{fig:qual4}
\end{figure*}

This section provides qualitative results of the video captioning experiment. We report the ground-truth captions annotated by humans, captions generated without using our localization module, as well as captions generated using our localization module.
Fig.~\ref{fig:qual4} shows that leveraging our localization module helps generate more descriptive and semantically correct captions, being able to describe the video better and hence assisting in the teleological understanding.

% \clearpage
% {
% \small
% \bibliographystyle{ieee_fullname}
% \bibliography{egbib}
% }

\clearpage

{
\small
\bibliographystyle{ieee_fullname}
\bibliography{egbib}

\begin{thebibliography}{10}\itemsep=-1pt

\bibitem{banerjee2005meteor}
Satanjeev Banerjee and Alon Lavie.
\newblock Meteor: An automatic metric for mt evaluation with improved
  correlation with human judgments.
\newblock In {\em Proceedings of the acl workshop on intrinsic and extrinsic
  evaluation measures for machine translation and/or summarization}, pages
  65--72, 2005.

\bibitem{brandone2009you}
Amanda~C Brandone and Henry~M Wellman.
\newblock You can't always get what you want: Infants understand failed
  goal-directed actions.
\newblock {\em Psychological science}, 20(1):85--91, 2009.

\bibitem{buch2017sst}
Shyamal Buch, Victor Escorcia, Chuanqi Shen, Bernard Ghanem, and Juan
  Carlos~Niebles.
\newblock Sst: Single-stream temporal action proposals.
\newblock In {\em Proceedings of the IEEE conference on Computer Vision and
  Pattern Recognition}, pages 2911--2920, 2017.

\bibitem{cao2019openpose}
Zhe Cao, Gines Hidalgo, Tomas Simon, Shih-En Wei, and Yaser Sheikh.
\newblock Openpose: realtime multi-person 2d pose estimation using part
  affinity fields.
\newblock {\em IEEE transactions on pattern analysis and machine intelligence},
  43(1):172--186, 2019.

\bibitem{transformer-cv2}
Nicolas Carion, Francisco Massa, Gabriel Synnaeve, Nicolas Usunier, Alexander
  Kirillov, and Sergey Zagoruyko.
\newblock End-to-end object detection with transformers.
\newblock In {\em European Conference on Computer Vision}, pages 213--229.
  Springer, 2020.

\bibitem{carreira2017quo}
Joao Carreira and Andrew Zisserman.
\newblock Quo vadis, action recognition? a new model and the kinetics dataset.
\newblock In {\em proceedings of the IEEE Conference on Computer Vision and
  Pattern Recognition}, pages 6299--6308, 2017.

\bibitem{chen2020look}
Zhenfang Chen, Lin Ma, Wenhan Luo, Peng Tang, and Kwan-Yee~K Wong.
\newblock Look closer to ground better: Weakly-supervised temporal grounding of
  sentence in video.
\newblock {\em arXiv preprint arXiv:2001.09308}, 2020.

\bibitem{chung2014empirical}
Junyoung Chung, Caglar Gulcehre, KyungHyun Cho, and Yoshua Bengio.
\newblock Empirical evaluation of gated recurrent neural networks on sequence
  modeling.
\newblock {\em arXiv preprint arXiv:1412.3555}, 2014.

\bibitem{csibra2008goal}
Gergely Csibra.
\newblock Goal attribution to inanimate agents by 6.5-month-old infants.
\newblock {\em Cognition}, 107(2):705--717, 2008.

\bibitem{csibra2007obsessed}
Gergely Csibra and Gy{\"o}rgy Gergely.
\newblock ‘obsessed with goals’: Functions and mechanisms of teleological
  interpretation of actions in humans.
\newblock {\em Acta psychologica}, 124(1):60--78, 2007.

\bibitem{damen2018scaling}
Dima Damen, Hazel Doughty, Giovanni~Maria Farinella, Sanja Fidler, Antonino
  Furnari, Evangelos Kazakos, Davide Moltisanti, Jonathan Munro, Toby Perrett,
  Will Price, et~al.
\newblock Scaling egocentric vision: The epic-kitchens dataset.
\newblock In {\em Proceedings of the European Conference on Computer Vision
  (ECCV)}, pages 720--736, 2018.

\bibitem{ar2}
Mahdi Davoodikakhki and KangKang Yin.
\newblock Hierarchical action classification with network pruning.
\newblock In {\em International Symposium on Visual Computing}, pages 291--305.
  Springer, 2020.

\bibitem{transformer-nlp1}
Jacob Devlin, Ming-Wei Chang, Kenton Lee, and Kristina Toutanova.
\newblock Bert: Pre-training of deep bidirectional transformers for language
  understanding.
\newblock {\em arXiv preprint arXiv:1810.04805}, 2018.

\bibitem{transformer-cv4}
Alexey Dosovitskiy, Lucas Beyer, Alexander Kolesnikov, Dirk Weissenborn,
  Xiaohua Zhai, Thomas Unterthiner, Mostafa Dehghani, Matthias Minderer, Georg
  Heigold, Sylvain Gelly, et~al.
\newblock An image is worth 16x16 words: Transformers for image recognition at
  scale.
\newblock {\em arXiv preprint arXiv:2010.11929}, 2020.

\bibitem{Epstein_2020_CVPR}
Dave Epstein, Boyuan Chen, and Carl Vondrick.
\newblock Oops! predicting unintentional action in video.
\newblock In {\em The IEEE/CVF Conference on Computer Vision and Pattern
  Recognition (CVPR)}, June 2020.

\bibitem{fang2020video2commonsense}
Zhiyuan Fang, Tejas Gokhale, Pratyay Banerjee, Chitta Baral, and Yezhou Yang.
\newblock Video2commonsense: Generating commonsense descriptions to enrich
  video captioning.
\newblock {\em Conference on Empirical Methods in Natural Language Processing},
  2020.

\bibitem{fang2019modularized}
Zhiyuan Fang, Shu Kong, Charless Fowlkes, and Yezhou Yang.
\newblock Modularized textual grounding for counterfactual resilience.
\newblock In {\em Proceedings of the IEEE/CVF Conference on Computer Vision and
  Pattern Recognition}, pages 6378--6388, 2019.

\bibitem{fang2020weak}
Zhiyuan Fang, Shu Kong, Zhe Wang, Charless Fowlkes, and Yezhou Yang.
\newblock Weak supervision and referring attention for temporal-textual
  association learning.
\newblock {\em arXiv preprint arXiv:2006.11747}, 2020.

\bibitem{fang2018weakly}
Zhiyuan Fang, Shu Kong, Tianshu Yu, and Yezhou Yang.
\newblock Weakly supervised attention learning for textual phrases grounding.
\newblock {\em arXiv preprint arXiv:1805.00545}, 2018.

\bibitem{fang2019intention}
Zhijie Fang and Antonio~M L{\'o}pez.
\newblock Intention recognition of pedestrians and cyclists by 2d pose
  estimation.
\newblock {\em IEEE Transactions on Intelligent Transportation Systems},
  21(11):4773--4783, 2019.

\bibitem{feinglass2021smurf}
Joshua Feinglass and Yezhou Yang.
\newblock Smurf: Semantic and linguistic understanding fusion for caption
  evaluation via typicality analysis.
\newblock In {\em Proceedings of the Annual Meeting of the Association for
  Computational Linguistics}, 2021.

\bibitem{rolling-unrolling}
Antonino Furnari and Giovanni~Maria Farinella.
\newblock What would you expect? anticipating egocentric actions with
  rolling-unrolling lstms and modality attention.
\newblock {\em CoRR}, abs/1905.09035, 2019.

\bibitem{gergely2003teleological}
Gy{\"o}rgy Gergely and Gergely Csibra.
\newblock Teleological reasoning in infancy: The na{\i}ve theory of rational
  action.
\newblock {\em Trends in cognitive sciences}, 7(7):287--292, 2003.

\bibitem{hoai2012max}
M. {Hoai} and F. {De la Torre}.
\newblock Max-margin early event detectors.
\newblock In {\em 2012 IEEE Conference on Computer Vision and Pattern
  Recognition}, pages 2863--2870, 2012.

\bibitem{fall2}
Minjie Hua, Yibing Nan, and Shiguo Lian.
\newblock Falls prediction based on body keypoints and seq2seq architecture.
\newblock In {\em Proceedings of the IEEE/CVF International Conference on
  Computer Vision Workshops}, pages 0--0, 2019.

\bibitem{ar1}
M~Esat Kalfaoglu, Sinan Kalkan, and A~Aydin Alatan.
\newblock Late temporal modeling in 3d cnn architectures with bert for action
  recognition.
\newblock In {\em European Conference on Computer Vision}, pages 731--747.
  Springer, 2020.

\bibitem{kalogeiton2017action}
Vicky Kalogeiton, Philippe Weinzaepfel, Vittorio Ferrari, and Cordelia Schmid.
\newblock Action tubelet detector for spatio-temporal action localization.
\newblock In {\em Proceedings of the IEEE International Conference on Computer
  Vision}, pages 4405--4413, 2017.

\bibitem{kay2017kinetics}
Will Kay, Joao Carreira, Karen Simonyan, Brian Zhang, Chloe Hillier, Sudheendra
  Vijayanarasimhan, Fabio Viola, Tim Green, Trevor Back, Paul Natsev, et~al.
\newblock The kinetics human action video dataset.
\newblock {\em arXiv preprint arXiv:1705.06950}, 2017.

\bibitem{transformer-nlp3}
Nitish~Shirish Keskar, Bryan McCann, Lav~R Varshney, Caiming Xiong, and Richard
  Socher.
\newblock Ctrl: A conditional transformer language model for controllable
  generation.
\newblock {\em arXiv preprint arXiv:1909.05858}, 2019.

\bibitem{kingma2014adam}
Diederik~P Kingma and Jimmy Ba.
\newblock Adam: A method for stochastic optimization.
\newblock {\em arXiv preprint arXiv:1412.6980}, 2014.

\bibitem{wsal7}
Pilhyeon Lee, Youngjung Uh, and Hyeran Byun.
\newblock Background suppression network for weakly-supervised temporal action
  localization.
\newblock In {\em Proceedings of the AAAI Conference on Artificial
  Intelligence}, volume~34, pages 11320--11327, 2020.

\bibitem{lei2020more}
Jie Lei, Licheng Yu, Tamara~L Berg, and Mohit Bansal.
\newblock What is more likely to happen next? video-and-language future event
  prediction.
\newblock {\em arXiv preprint arXiv:2010.07999}, 2020.

\bibitem{li2019angular}
Zhaoqun Li, Cheng Xu, and Biao Leng.
\newblock Angular triplet-center loss for multi-view 3d shape retrieval.
\newblock In {\em Proceedings of the AAAI Conference on Artificial
  Intelligence}, volume~33, pages 8682--8689, 2019.

\bibitem{lin2004rouge}
Chin-Yew Lin.
\newblock Rouge: A package for automatic evaluation of summaries.
\newblock In {\em Text summarization branches out}, pages 74--81, 2004.

\bibitem{ar5}
Ji Lin, Chuang Gan, and Song Han.
\newblock Tsm: Temporal shift module for efficient video understanding.
\newblock In {\em Proceedings of the IEEE/CVF International Conference on
  Computer Vision}, pages 7083--7093, 2019.

\bibitem{wsal5}
Daochang Liu, Tingting Jiang, and Yizhou Wang.
\newblock Completeness modeling and context separation for weakly supervised
  temporal action localization.
\newblock In {\em Proceedings of the IEEE/CVF Conference on Computer Vision and
  Pattern Recognition}, pages 1298--1307, 2019.

\bibitem{pose1}
Diogo~C Luvizon, David Picard, and Hedi Tabia.
\newblock 2d/3d pose estimation and action recognition using multitask deep
  learning.
\newblock In {\em Proceedings of the IEEE Conference on Computer Vision and
  Pattern Recognition}, pages 5137--5146, 2018.

\bibitem{ma2020vlanet}
Minuk Ma, Sunjae Yoon, Junyeong Kim, Youngjoon Lee, Sunghun Kang, and Chang~D
  Yoo.
\newblock Vlanet: Video-language alignment network for weakly-supervised video
  moment retrieval.
\newblock In {\em European Conference on Computer Vision}, pages 156--171.
  Springer, 2020.

\bibitem{miech2019leveraging}
A. {Miech}, I. {Laptev}, J. {Sivic}, H. {Wang}, L. {Torresani}, and D. {Tran}.
\newblock Leveraging the present to anticipate the future in videos.
\newblock In {\em 2019 IEEE/CVF Conference on Computer Vision and Pattern
  Recognition Workshops (CVPRW)}, pages 2915--2922, 2019.

\bibitem{skeleton-loc1}
Daisuke Miki, Shi Chen, and Kazuyuki Demachi.
\newblock Weakly supervised graph convolutional neural network for human action
  localization.
\newblock In {\em Proceedings of the IEEE/CVF Winter Conference on Applications
  of Computer Vision}, pages 653--661, 2020.

\bibitem{wsal2}
Kyle Min and Jason~J Corso.
\newblock Adversarial background-aware loss for weakly-supervised temporal
  activity localization.
\newblock In {\em European Conference on Computer Vision}, pages 283--299.
  Springer, 2020.

\bibitem{Mithun_2019_CVPR}
Niluthpol~Chowdhury Mithun, Sujoy Paul, and Amit~K. Roy-Chowdhury.
\newblock Weakly supervised video moment retrieval from text queries.
\newblock In {\em The IEEE Conference on Computer Vision and Pattern
  Recognition (CVPR)}, June 2019.

\bibitem{nguyen2018weakly}
Phuc Nguyen, Ting Liu, Gautam Prasad, and Bohyung Han.
\newblock Weakly supervised action localization by sparse temporal pooling
  network.
\newblock In {\em Proceedings of the IEEE Conference on Computer Vision and
  Pattern Recognition}, pages 6752--6761, 2018.

\bibitem{wsal6}
Sujoy Paul, Sourya Roy, and Amit~K Roy-Chowdhury.
\newblock W-talc: Weakly-supervised temporal activity localization and
  classification.
\newblock In {\em Proceedings of the European Conference on Computer Vision
  (ECCV)}, pages 563--579, 2018.

\bibitem{paul2018w}
Sujoy Paul, Sourya Roy, and Amit~K Roy-Chowdhury.
\newblock W-talc: Weakly-supervised temporal activity localization and
  classification.
\newblock In {\em Proceedings of the European Conference on Computer Vision
  (ECCV)}, pages 563--579, 2018.

\bibitem{JAAD}
A. {Rasouli}, I. {Kotseruba}, and J.~K. {Tsotsos}.
\newblock Are they going to cross? a benchmark dataset and baseline for
  pedestrian crosswalk behavior.
\newblock In {\em 2017 IEEE International Conference on Computer Vision
  Workshops (ICCVW)}, pages 206--213, 2017.

\bibitem{rasouli2020pedestrian}
Amir Rasouli, Mohsen Rohani, and Jun Luo.
\newblock Pedestrian behavior prediction via multitask learning and categorical
  interaction modeling.
\newblock {\em arXiv preprint arXiv:2012.03298}, 2020.

\bibitem{ryoo2011early}
M.~S. {Ryoo}.
\newblock Human activity prediction: Early recognition of ongoing activities
  from streaming videos.
\newblock In {\em 2011 International Conference on Computer Vision}, pages
  1036--1043, 2011.

\bibitem{sadegh2017encouraging}
Mohammad Sadegh~Aliakbarian, Fatemeh Sadat~Saleh, Mathieu Salzmann, Basura
  Fernando, Lars Petersson, and Lars Andersson.
\newblock Encouraging lstms to anticipate actions very early.
\newblock In {\em Proceedings of the IEEE International Conference on Computer
  Vision}, pages 280--289, 2017.

\bibitem{wsal3}
Baifeng Shi, Qi Dai, Yadong Mu, and Jingdong Wang.
\newblock Weakly-supervised action localization by generative attention
  modeling.
\newblock In {\em Proceedings of the IEEE/CVF Conference on Computer Vision and
  Pattern Recognition}, pages 1009--1019, 2020.

\bibitem{shou2017cdc}
Zheng Shou, Jonathan Chan, Alireza Zareian, Kazuyuki Miyazawa, and Shih-Fu
  Chang.
\newblock Cdc: Convolutional-de-convolutional networks for precise temporal
  action localization in untrimmed videos.
\newblock In {\em Proceedings of the IEEE Conference on Computer Vision and
  Pattern Recognition}, pages 5734--5743, 2017.

\bibitem{shou2018autoloc}
Zheng Shou, Hang Gao, Lei Zhang, Kazuyuki Miyazawa, and Shih-Fu Chang.
\newblock Autoloc: Weakly-supervised temporal action localization in untrimmed
  videos.
\newblock In {\em Proceedings of the European Conference on Computer Vision
  (ECCV)}, pages 154--171, 2018.

\bibitem{shou2016temporal}
Zheng Shou, Dongang Wang, and Shih-Fu Chang.
\newblock Temporal action localization in untrimmed videos via multi-stage
  cnns.
\newblock In {\em Proceedings of the IEEE Conference on Computer Vision and
  Pattern Recognition}, pages 1049--1058, 2016.

\bibitem{fall1}
Markus~D Solbach and John~K Tsotsos.
\newblock Vision-based fallen person detection for the elderly.
\newblock In {\em Proceedings of the IEEE International Conference on Computer
  Vision Workshops}, pages 1433--1442, 2017.

\bibitem{sommerville2005pulling}
Jessica~A Sommerville and Amanda~L Woodward.
\newblock Pulling out the intentional structure of action: the relation between
  action processing and action production in infancy.
\newblock {\em Cognition}, 95(1):1--30, 2005.

\bibitem{song2020weakly}
Yijun Song, Jingwen Wang, Lin Ma, Zhou Yu, and Jun Yu.
\newblock Weakly-supervised multi-level attentional reconstruction network for
  grounding textual queries in videos.
\newblock {\em arXiv preprint arXiv:2003.07048}, 2020.

\bibitem{sun2015temporal}
Chen Sun, Sanketh Shetty, Rahul Sukthankar, and Ram Nevatia.
\newblock Temporal localization of fine-grained actions in videos by domain
  transfer from web images.
\newblock In {\em Proceedings of the 23rd ACM international conference on
  Multimedia}, pages 371--380. ACM, 2015.

\bibitem{Synakowski_2021}
Stuart Synakowski, Qianli Feng, and Aleix Martinez.
\newblock Adding knowledge to unsupervised algorithms for the recognition of
  intent.
\newblock {\em International Journal of Computer Vision}, Jan 2021.

\bibitem{tan2020learning}
Ganchao Tan, Daqing Liu, Meng Wang, and Zheng-Jun Zha.
\newblock Learning to discretely compose reasoning module networks for video
  captioning.
\newblock {\em arXiv preprint arXiv:2007.09049}, 2020.

\bibitem{tran2018closer}
Du Tran, Heng Wang, Lorenzo Torresani, Jamie Ray, Yann LeCun, and Manohar
  Paluri.
\newblock A closer look at spatiotemporal convolutions for action recognition.
\newblock In {\em Proceedings of the IEEE conference on Computer Vision and
  Pattern Recognition}, pages 6450--6459, 2018.

\bibitem{tran2012max}
Du Tran and Junsong Yuan.
\newblock Max-margin structured output regression for spatio-temporal action
  localization.
\newblock In {\em Advances in neural information processing systems}, pages
  350--358, 2012.

\bibitem{back-to-future}
Vinh Tran, Yang Wang, and Minh Hoai.
\newblock Back to the future: Knowledge distillation for human action
  anticipation.
\newblock {\em CoRR}, abs/1904.04868, 2019.

\bibitem{intent-urban}
D. {Varytimidis}, F. {Alonso-Fernandez}, B. {Duran}, and C. {Englund}.
\newblock Action and intention recognition of pedestrians in urban traffic.
\newblock In {\em 2018 14th International Conference on Signal-Image Technology
  Internet-Based Systems (SITIS)}, pages 676--682, 2018.

\bibitem{NIPS2017_3f5ee243}
Ashish Vaswani, Noam Shazeer, Niki Parmar, Jakob Uszkoreit, Llion Jones,
  Aidan~N Gomez, \L~ukasz Kaiser, and Illia Polosukhin.
\newblock Attention is all you need.
\newblock In I. Guyon, U.~V. Luxburg, S. Bengio, H. Wallach, R. Fergus, S.
  Vishwanathan, and R. Garnett, editors, {\em Advances in Neural Information
  Processing Systems}, volume~30. Curran Associates, Inc., 2017.

\bibitem{vedantam2015cider}
Ramakrishna Vedantam, C Lawrence~Zitnick, and Devi Parikh.
\newblock Cider: Consensus-based image description evaluation.
\newblock In {\em Proceedings of the IEEE conference on computer vision and
  pattern recognition}, pages 4566--4575, 2015.

\bibitem{vondrick2016predicting}
Carl Vondrick, Deniz Oktay, Hamed Pirsiavash, and Antonio Torralba.
\newblock Predicting motivations of actions by leveraging text.
\newblock In {\em Proceedings of the IEEE conference on computer vision and
  pattern recognition}, pages 2997--3005, 2016.

\bibitem{rep-pred}
Carl Vondrick, Hamed Pirsiavash, and Antonio Torralba.
\newblock Anticipating the future by watching unlabeled video.
\newblock {\em CoRR}, abs/1504.08023, 2015.

\bibitem{pose2}
Chunyu Wang, Yizhou Wang, and Alan~L Yuille.
\newblock An approach to pose-based action recognition.
\newblock In {\em Proceedings of the IEEE conference on computer vision and
  pattern recognition}, pages 915--922, 2013.

\bibitem{ar3}
Limin Wang, Yuanjun Xiong, Zhe Wang, Yu Qiao, Dahua Lin, Xiaoou Tang, and Luc
  Van~Gool.
\newblock Temporal segment networks: Towards good practices for deep action
  recognition.
\newblock In {\em European conference on computer vision}, pages 20--36.
  Springer, 2016.

\bibitem{where-why}
P. {Wei}, Y. {Liu}, T. {Shu}, N. {Zheng}, and S. {Zhu}.
\newblock Where and why are they looking? jointly inferring human attention and
  intentions in complex tasks.
\newblock In {\em 2018 IEEE/CVF Conference on Computer Vision and Pattern
  Recognition}, pages 6801--6809, 2018.

\bibitem{weinzaepfel2015learning}
Philippe Weinzaepfel, Zaid Harchaoui, and Cordelia Schmid.
\newblock Learning to track for spatio-temporal action localization.
\newblock In {\em Proceedings of the IEEE international conference on computer
  vision}, pages 3164--3172, 2015.

\bibitem{wojke2017simple}
Nicolai Wojke, Alex Bewley, and Dietrich Paulus.
\newblock Simple online and realtime tracking with a deep association metric.
\newblock In {\em 2017 IEEE international conference on image processing
  (ICIP)}, pages 3645--3649. IEEE, 2017.

\bibitem{woodward2001infants}
Amanda~L Woodward, Jessica~A Sommerville, and Jose~J Guajardo.
\newblock How infants make sense of intentional action.
\newblock {\em Intentions and intentionality: Foundations of social cognition},
  pages 149--169, 2001.

\bibitem{transformer-nlp4}
Chien-Sheng Wu, Steven Hoi, Richard Socher, and Caiming Xiong.
\newblock Tod-bert: Pre-trained natural language understanding for
  task-oriented dialogues.
\newblock {\em arXiv preprint arXiv:2004.06871}, 2020.

\bibitem{pose3}
Sijie Yan, Yuanjun Xiong, and Dahua Lin.
\newblock Spatial temporal graph convolutional networks for skeleton-based
  action recognition.
\newblock In {\em Proceedings of the AAAI conference on artificial
  intelligence}, volume~32, 2018.

\bibitem{transformer-nlp2}
Zhilin Yang, Zihang Dai, Yiming Yang, Jaime Carbonell, Ruslan Salakhutdinov,
  and Quoc~V Le.
\newblock Xlnet: Generalized autoregressive pretraining for language
  understanding.
\newblock {\em arXiv preprint arXiv:1906.08237}, 2019.

\bibitem{zellers2019vcr}
Rowan Zellers, Yonatan Bisk, Ali Farhadi, and Yejin Choi.
\newblock From recognition to cognition: Visual commonsense reasoning.
\newblock In {\em The IEEE Conference on Computer Vision and Pattern
  Recognition (CVPR)}, June 2019.

\bibitem{transformer-cv3}
Yanhong Zeng, Jianlong Fu, and Hongyang Chao.
\newblock Learning joint spatial-temporal transformations for video inpainting.
\newblock In {\em European Conference on Computer Vision}, pages 528--543.
  Springer, 2020.

\bibitem{wsal4}
Yuanhao Zhai, Le Wang, Wei Tang, Qilin Zhang, Junsong Yuan, and Gang Hua.
\newblock Two-stream consensus network for weakly-supervised temporal action
  localization.
\newblock In {\em European Conference on Computer Vision}, pages 37--54.
  Springer, 2020.

\bibitem{transformer-cv1}
Dong Zhang, Hanwang Zhang, Jinhui Tang, Meng Wang, Xiansheng Hua, and Qianru
  Sun.
\newblock Feature pyramid transformer.
\newblock In {\em European Conference on Computer Vision}, pages 323--339.
  Springer, 2020.

\bibitem{transformer-cv6}
Ziqi Zhang, Yaya Shi, Chunfeng Yuan, Bing Li, Peijin Wang, Weiming Hu, and
  Zheng-Jun Zha.
\newblock Object relational graph with teacher-recommended learning for video
  captioning.
\newblock In {\em Proceedings of the IEEE/CVF conference on computer vision and
  pattern recognition}, pages 13278--13288, 2020.

\bibitem{ar4}
Bolei Zhou, Alex Andonian, Aude Oliva, and Antonio Torralba.
\newblock Temporal relational reasoning in videos.
\newblock In {\em Proceedings of the European Conference on Computer Vision
  (ECCV)}, pages 803--818, 2018.

\bibitem{transformer-cv5}
Luowei Zhou, Yingbo Zhou, Jason~J Corso, Richard Socher, and Caiming Xiong.
\newblock End-to-end dense video captioning with masked transformer.
\newblock In {\em Proceedings of the IEEE Conference on Computer Vision and
  Pattern Recognition}, pages 8739--8748, 2018.

\bibitem{zhou2004multi}
Zhi-Hua Zhou.
\newblock Multi-instance learning: A survey.
\newblock {\em Department of Computer Science \& Technology, Nanjing
  University, Tech. Rep}, 2, 2004.

\end{thebibliography}
}

\end{document}